\documentclass{article}

\usepackage{microtype}
\usepackage{graphicx}
\usepackage{caption}
\usepackage{subcaption}
\usepackage{array}
\usepackage{booktabs} 
\usepackage{soul}
\usepackage{makecell}
\usepackage{dblfloatfix}

\usepackage{hyperref}

\usepackage[accepted]{icml2024}

\usepackage{amsmath,amsfonts,bm}

\def\eqref#1{equation~\ref{#1}}

\def\1{\bm{1}}

\def\vc{{\bm{c}}}

\def\vz{{\bm{z}}}

\DeclareMathAlphabet{\mathsfit}{\encodingdefault}{\sfdefault}{m}{sl}
\SetMathAlphabet{\mathsfit}{bold}{\encodingdefault}{\sfdefault}{bx}{n}

\DeclareMathOperator*{\argmax}{arg\,max}

\usepackage{amsmath}
\usepackage{amssymb}
\usepackage{mathtools}
\usepackage{amsthm}

\usepackage[capitalize,noabbrev]{cleveref}

\theoremstyle{plain}

\theoremstyle{definition}

\theoremstyle{remark}

\usepackage[textsize=tiny]{todonotes}

\icmltitlerunning{Estimating Unknown Population Sizes Using the Hypergeometric Distribution}

\begin{document}

\twocolumn[
\icmltitle{Estimating Unknown Population Sizes Using the Hypergeometric Distribution}



\icmlsetsymbol{equal}{*}

\begin{icmlauthorlist}
\icmlauthor{Liam Hodgson}{mcgill,mila}
\icmlauthor{Danilo Bzdok}{mcgill,mila}
\end{icmlauthorlist}

\icmlaffiliation{mcgill}{McGill University, Montréal, Canada}
\icmlaffiliation{mila}{Mila - Québec Artificial Intelligence Institute}

\icmlcorrespondingauthor{Danilo Bzdok}{danilobzdok@gmail.com}

\icmlkeywords{Machine Learning, ICML}

\vskip 0.3in
]



\printAffiliationsAndNotice{} 

\begin{abstract}
The multivariate hypergeometric distribution describes sampling without replacement from a discrete population of elements divided into multiple categories.
Addressing a gap in the literature, we tackle the challenge of estimating discrete distributions when both the total population size and the category sizes are unknown.
Here, we propose a novel solution using the hypergeometric likelihood to solve this estimation problem, even in the presence of severe under-sampling.
Our approach accounts for a data generating process where the ground-truth is a mixture of distributions conditional on a continuous latent variable, as seen in collaborative filtering, using the variational autoencoder framework.
Empirical data simulation demonstrates that our method outperforms other likelihood functions used to model count data, both in terms of accuracy of population size estimate and learning an informative latent space.
We showcase our method's versatility through applications in NLP, by inferring and estimating the complexity of latent vocabularies in reading passage excerpts, and in biology, by accurately recovering the true number of gene transcripts from sparse single-cell genomics data.
\end{abstract}

\section{Introduction}
The classic Pólya urn model \citep{eggenberger_uber_1923} describes the process of randomly sampling from an urn containing balls of various colors, and is used to illustrate common discrete probability distributions that form the core building block of many probabilistic machine learning models. When balls are sampled from the urn \textit{with} replacement, the distribution of counts of balls of each color is described by the multinomial distribution, whereas the hypergeometric distribution describes sampling \textit{without} replacement. The hypergeometric distribution becomes important for successful modelling when the selection of one element from the distribution affects the probabilities of selecting subsequent elements (i.e. the counts within as sample are not independent), and when the number of elements sampled is significant compared to the population size. The hypergeometric distribution also enables the direct inference of category counts, as opposed to category probabilities, which is often of interest for downstream analyses and interpretability.

There are many settings where it is valuable to model count data directly and to capture the dependence between category counts. For example, in the context of collaborative filtering, song or movie play counts \cite{van2013deep} and shopping basket item counts \cite{faggioli2020recency} can be thought of as being sampled from a larger catalogue. Likewise, a text document can be viewed as a bag-of-words that is sampled from a larger, underlying bag-of-words \cite{goldberg2022neural}. In all of these cases the assumption of sampling without replacement is justifiable because the magnitude of counts is relatively small, and there is dependence between these counts. For example, watching a movie changes the probability of it being watched a second time, adding an item to a virtual shopping basket changes the probability of it being added again, and selecting one word changes the probability of its synonyms and related terms also appearing in a text.

This setting is even more common in biology due to the experimental sampling methods used \cite{depatta1998sampling}. In the field of single-cell genomics, the quantity of gene transcript count data measured at the resolution of individual cells is accumulating at an exponential rate \cite{svensson2018exponential}. As we show in Section \ref{sec:sc}, high-throughput experiments inherently sample without replacement, and with careful interpretation this single-cell data promises to answer important questions in biology and health. In the field of microbiome research, another potential application is the estimation of microbial population sizes and subsequent differential abundance analysis \citep{morton2019learning,morton2019establishing}. Due to the frequent occurrence of finite, discrete populations in nature, there are many other natural phenomena that can be modelled using the hypergeometric distribution \cite{holmes2018modern}.

Furthermore, the aforementioned applications are typically characterized by intrinsic low-rank structure. Movie and music choices are driven by an underlying set of preferences \cite{feuerverger2012statistical}, words in a text passage are associated with the document's topic \cite{vayansky2020review}, and a cell’s gene transcript counts depend strongly on its cell type \cite{gronbech2020scvae}. This latent structure induces correlation between features, such as songs of the same genre, and suggests that a latent-factor model is appropriate.

Despite the fundamental importance of the hypergeometric distribution - it being one of the few fundamental probability distributions for modeling sampling - there is currently no effective way of inferring its parameters in many common settings of interest, including in the presence of high-dimensional data with intrinsic low-rank structure. In this paper, we present a simple yet powerful method to solve an as yet unaddressed core problem for the hypergeometric distribution, namely inferring a mixture of discrete distributions where both the size of the total population and its constituent categories are unknown. We first show that parameter inference is tractable when there are two or more categories using empirical simulations, and that our method outperforms existing likelihood-based approaches. To demonstrate its value in real-world settings, we use our method to infer a latent vocabulary for reading passages, which can be used to assess their complexity. Finally, we address the technical limitations of genomics experimental methods and recover missing gene transcript counts in high-throughput single-cell genomics measurements.

\section{Background and Related Methods}
\subsection{The Hypergeometric Distribution}
Consider an urn that contains $N$ balls divided into $K=2$ categories (e.g. colors): $N_1$ balls are white and $N_2 = N - N_1$ are black. Each ball has equal probability of being selected from the urn. If we sample $n<N$ balls \textit{without replacement}, obtaining $c_1$ white ball counts and $c_2$ black counts, the distribution of the number of sampled balls of each color is given by the hypergeometric distribution, whose joint probability mass function is \citep{Moivre1711}:
\begin{align}
    P(c_1,c_2|N_1,N_2) = \frac{\binom{N_1}{c_1} \binom{N_2}{c_2}}{\binom{N_1+N_2}{c_1+c_2}}
\end{align}

In general, when we have $K$ categories the joint probability mass function is:
\begin{align}
    P(c_1,\ldots,c_K|N_1,\ldots,N_K) = \frac{\prod_{i=1}^K \binom{N_i}{c_i}}{\binom{\sum_{i=1}^K N_i}{\sum_{i=1}^K c_i}}
\end{align}

Typically $N$ is assumed to be known, however here we consider the setting where neither $N$ nor any $N_i$ are known. In the typical setting the distribution for $K=2$ is called the univariate hypergeometric distribution, with $K>2$ referred to as multivariate, but because in our problem setting there are already two unknown variables when $K=2$, we use the term hypergeometric distribution for any $K \ge 2$.

\subsection{Existing Maximum Likelihood Estimators} 
There are two standard maximum likelihood estimation problems that have been investigated for the hypergeometric distribution.

\textit{Known total population size:} When the total population size $N$ is known, the object is to estimate the true number of elements $N_i$ in each constituent category $i \in 1,\ldots,K$. The maximum likelihood estimator is then essentially the known total population size scaled by the sample frequency of each category, with adjustment to ensure a correct integer solution \citep{oberhofer_maximum_1987}.

\textit{Unknown total population size}: In a more complex case known as the capture-recapture problem \citep{capture-recapture-1958}, the total population size $N$ is unknown. To estimate the total population, a sample is first drawn from the underlying distribution. All objects belonging to one of the categories in this sample are tagged, and all sampled elements are returned to the population. A second sample from the same distribution is then drawn, and the number of tagged samples that reappear permits the estimation of the total population size using maximum likelihood. This method has found important applications in biology and ecology. However, it depends on the ability to tag and resample the same population, which is often not possible in practice, and does not extend to a mixture of distributions.

Additionally, \citet{tohma_estimation_1991} propose a number of approximate methods for the estimation of the parameters of the hypergeometric distribution when the category sizes are unknown, including an approximation using the likelihood, however they do not maximize the likelihood directly.

\subsection{Related Methods}

The hypergeometric likelihood has for the most part been neglected in the modern machine learning context.
\citet{sutter_learning_2022} proposed a continuous relaxation of the non-central hypergeometric distribution to allow for differentiable sampling using the Gumbel-Softmax trick, and use it to learn category weights. This method assumes that the number of elements in each category are known, and therefore that only the category weights are to be estimated. \citet{waudby-smith_confidence_2020} presented a method for uncertainty quantification when sequentially sampling without replacement from a finite population, as defined by the hypergeometric distribution, to estimate confidence bounds as new data becomes available. Other parametric distributional assumptions can also be used to directly model count data, such as the Poisson and negative binomial, and a wide range of methods have been developed making these explicit parametric assumptions.

The multinomial distribution, which is the limiting form of the hypergeometric distribution when the sample size is negligible relative to the population ($N>>n$), appears frequently in machine learning literature. For example, it is often used to model counts based on their relative abundance. Latent Dirichlet allocation \citep{lda_blei_2003} uses a hierarchical generative framework to model the distribution of topics, documents over topics, and word (category) counts from a vocabulary (population) over documents. The distributions over topics and words are multinomial, with word counts being transformed into their relative frequency in the document.
\cite{liang_variational_2018} use a variational autoencoder with multinomial likelihood applied to collaborative filtering for recommender systems. The click data and play counts are binarized to accommodate the multinomial likelihood. 

\citet{awasthi2021benefits} argue for the use of maximum likelihood estimation instead of empirical risk minimization, showing that it is better at capturing the appropriate inductive bias and that its performance is competitive with direct minimization of a target metric.

\section{Method}
\subsection{Single Ground-Truth Distribution}
We first assume that there exists a single true discrete population of elements, with total number of elements $N$ belonging to $K$ categories. We consider the setting where none of the category sizes are known, and we wish to estimate them from the data; that is, we do not know any category size $N_i$, nor the total population size $N = \sum_{i=1}^K N_i$. We assume under-sampling of the underlying distribution ($n << N$) as this is typical real-world applications.

We consider $T$ independent trials, each producing an observed count vector $\vc_t = \{c_{t,1},\ldots,c_{t,K}\}$, $t \in 1,\ldots,T$. In trial $t \in T$ we draw $n_t$ samples, without replacement, from the discrete underlying population, such that $\sum_{i=1}^K c_{t,i} = n_t$. The likelihood for the hypergeometric distribution is:
\begin{align}
    \mathcal{L}(N_1,\ldots,N_K | \vc_1,\ldots,\vc_T) = \prod_{t=1}^T \frac{\prod_{i=1}^K \binom{N_i}{c_{t,i}}}{\binom{\sum_{i=1}^K N_i}{\sum_{i=1}^K c_{t,i}}}
\end{align}

The log-likelihood is:
\begin{align}
    \begin{split}
        &\log \mathcal{L}(N_1,\ldots,N_K | \vc_1,\ldots,\vc_T) = \\
        &\sum_{t=1}^T \left[ \sum_{i=1}^K \log \binom{N_i}{c_{t,i}} - \log \binom{\sum_{i=1}^K N_i}{\sum_{i=1}^K c_{t,i}} \right]
    \end{split}
\end{align}

The hypergeometric distribution is not part of the exponential family, so it is not obvious that a closed-form maximum likelihood estimator exists, and therefore we turn to numerical optimization methods. To enable continuous optimization, we consider a continuous and differentiable relaxation of the log-likelihood by replacing the factorials in the binomial coefficient with the gamma function, which is the extension of the factorial to real-numbered arguments:
\begin{align}
    \binom{a}{b} &= \frac{a!}{b!(a-b)!}
    = \frac{\Gamma(a+1)}{\Gamma(b+1)\Gamma(a-b+1)}
\end{align}

Using the resulting log-likelihood for the parameter set $\theta = \{ N_1,\ldots,N_K \}$, we perform maximum likelihood estimation to obtain the MLE $\displaystyle \hat{\theta} \equiv \{ \hat{N}_1,\ldots,\hat{N}_K \} = \argmax_\theta \log \mathcal{L}(\theta | \vc_1,\ldots,\vc_T)$. During training, the continuous estimates of $N_i$ are used to perform gradient updates, whereas during inference the estimates can be rounded to the nearest integer if a discrete solution is required.

The binomial coefficient is not defined if $c_k > \hat{N}_k$, which would correspond to the impossible scenario of sampling more balls of a given color than are present in the urn. To impose the requirement $\hat{N}_k \ge c_k$, we add a violation penalty $C_\text{viol}$ to the negative log-likelihood we seek to minimize (Equation \ref{math:violation1}), and we threshold any estimates of $\hat{N}_k < c_k$ at $c_k$ before evaluating the likelihood (Equation \ref{math:violation2}). The use of a violation function is a common approach to solving constrained optimization problems \cite{bertsekas2014constrained}. We threshold $N_i$ at the observed sample value $c_{t,i}$ as opposed to the minimum across all samples $\min_t c_{t,i}$ to remain as general as possible. This is because in the case of a mixture of distributions we do not know which observation originates from which underlying distribution, and hence do not know what the correct minimum is.
\begin{align}
    C_\text{viol} &= \sum_{i=1}^K \max(0, c_i - \hat{N}_i) \label{math:violation1} \\
    \hat{N}_i &\leftarrow \max(c_i, \hat{N}_i) \quad i \in 1, \ldots,K \label{math:violation2}
\end{align}

In this paper we are interested in modelling scenarios where we have access to many samples drawn from the same underlying population, but where this population is under-sampled. Specifically, if $N$ is the total population size, we assume that we obtain samples with at most $n_{max}$ objects drawn from the ground-truth distribution in each trial, giving a sample fraction $f_{max} = n_{max}/N$. For example, if $f_{max}=0.5$ and the true distribution has 100 elements, no observation will contain more than 50 samples.

\subsection{Mixture of Ground-Truth Distributions}
While the hypergeometric likelihood can be used to directly estimate the ground truth for a single underlying distribution using maximum likelihood (Section \ref{sec:single_dist}), data generated by real world processes is often better represented by a mixture of distributions. Therefore, we next propose a generative latent variable model $p(\vc,\vz) = p(\vc|\vz)p(\vz)$ to jointly model the observed counts conditional on a continuous latent variable $\vz$.

We extend our approach to allow for the estimation of a high-dimensional distribution that is conditional on a continuous latent variable, allowing it to capture a continuous mixture of count distributions. This modelling assumption is essential when the true distribution is considerably undersampled, leading to sparsity, and so information needs to be shared between similar observations to successfully model the data. The variational autoencoder (VAE) \citep{kingma_auto-encoding_2014} is a powerful framework for performing efficient estimation of the generative distribution parameters in the presence of a continuous latent variable and high-dimensional data.

Following the VAE framework, we assume a data generating process where a latent variable $\vz$ is first drawn from a prior distribution $p(\vz)$, then a count vector $\vc$ is generated from the conditional hypergeometric likelihood $p(\vc|\vz)$. Note again that because we are using the continuous relaxation of the hypergeometric distribution, the generated $\vc$ are continuous. The marginal hypergeometric likelihood that we are interested in maximizing, $p(\vc) = \int p(\vc|\vz) p(\vz) d\vz$, is intractable. We approximate the true unknown posterior for the latent variable $p(\vz|\vc)$ with the variational distribution $q_\phi(\vz|\vc)$, parameterized by a deep feed-forward neural network with parameters $\phi$. Similarly, we represent the conditional hypergeometric likelihood $p_\psi(\vc|\vz)$ with another neural network parameterized by $\psi$. We choose a factorized multivariate Gaussian (diagonal covariance) as the prior $p(z)$ over the latent variable. We optimize the variational lower bound, also known as the Evidence Lower Bound (ELBO), regularized by the violation penalty:
\begin{align}
    \begin{split}
        \mathcal{L}(\psi, \phi; \vc_t) = &-D_\text{KL}(q_\phi(\vz|\vc_t) || p(\vz))
        + \mathbb{E}_{q_\phi}\left[ \log p_\psi (\vc_t|\vz) \right] \\
        &+ \sum_{i=1}^K \max(0, c_i - \hat{N}_i)
    \end{split}
\end{align}
The first term is the KL divergence between the approximate posterior and the prior, the second term is the expected log-likelihood under the approximate posterior, and the third term is the violation penalty. Optimization of this objective enables the estimation of counts under-sampled from a mixture of distributions. The MLE estimate for $\theta_i$ is obtained by first computing $\vz_i = f(\vc_i)$, where $\vc_i$ is the observed count vector, then computing $\theta_i = g(\vz_i)$, where the output layer of $g$ is a linear layer with $K$ features and a ReLU activation.

\section{Experiments}
\subsection{Tractability of Inferring Ground-Truth Distributions}
\label{sec:single_dist}
Using empirical data simulation, we begin by demonstrating that this inference problem is tractable in the ideal case where observations are sampled from a single discrete ground-truth distribution. Despite its simplicity, this scenario has yet to have been shown tractable, and in fact appears to have generally been assumed to be intractable.

To simulate data, we first select the true number of objects $N_i$ in each category $k \in K$. We then generate observations from $T$ independent trials, where in trial $t \in 1,\ldots,T$ we sample $n_t$ objects from the true distribution, without replacement. For each trial, we first determine the extent of under-sampling by choosing the number of objects to draw $n_t \sim \text{Uniform}(2,n_{max})$, where $n_{max} < N$ is the maximum number of objects that can be sampled in any trial.

\begin{figure}[h!]
    \centering
    \includegraphics[width=0.8\columnwidth]{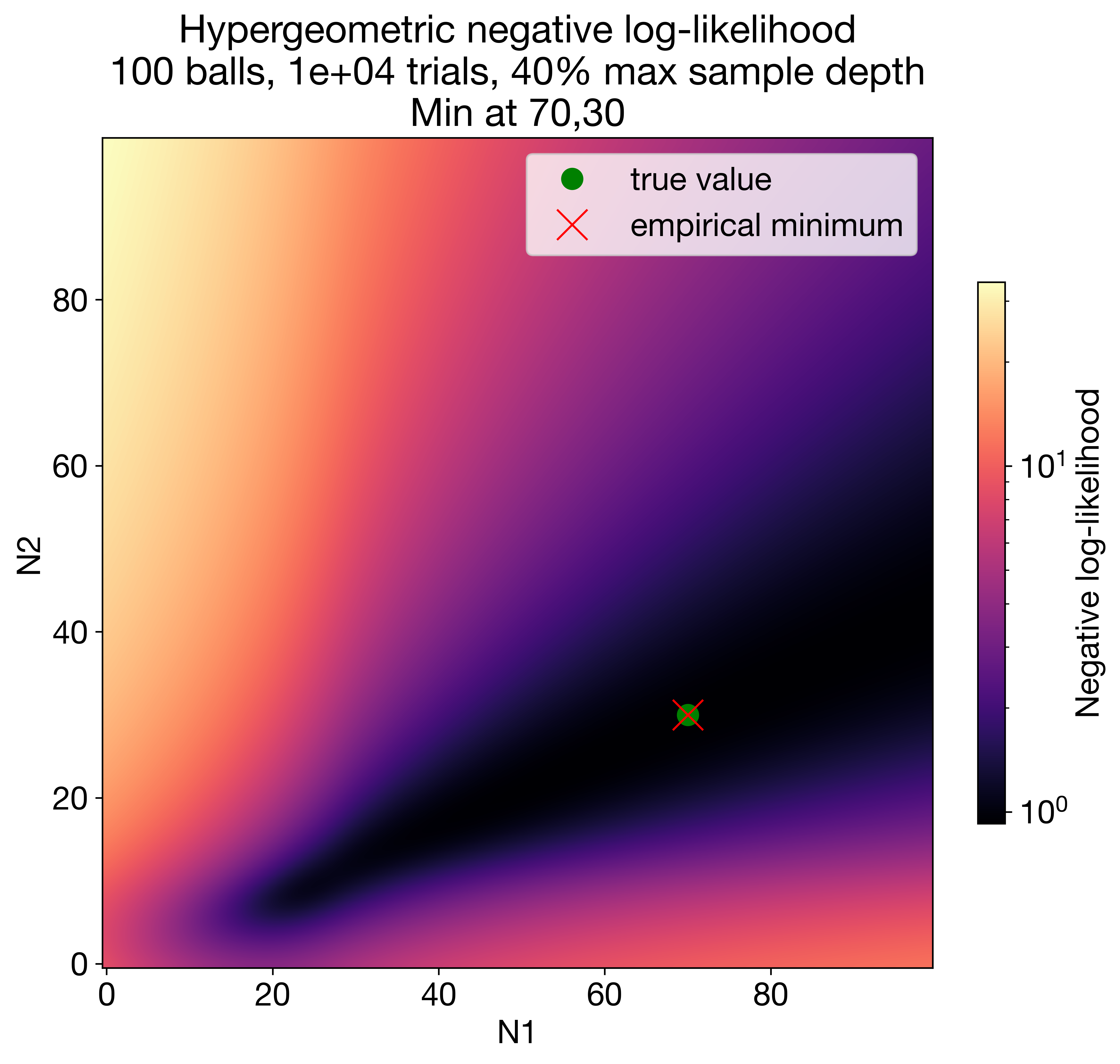}
    \caption{Example of negative log-likelihood landscape for $K=2$ and $N=100$ ($10^4$ trials).}
    \label{fig:landscape}
\end{figure}

To demonstrate the typical behavior, we show results for a simulation for the scenario where $N_1=70$ and $N_2=30$ ($N=100$, $K=2$, $f_{max}=0.4$), evaluating the negative log-likelihood (NLL) for all possible combinations of $\hat{N}_1$ and $\hat{N}_2$. Figure \ref{fig:landscape} shows the resulting loss landscape, and we see that the minimum NLL value corresponds to the true $N_1$ and $N_2$ values. The lowest NLL region lies along the line of correct $N_2/N_1$ ratio. We note that the difference in NLL between the optimum and surrounding estimates is small, especially for estimates with the same ratio $N_2/N_1$.

\begin{figure}[h!]
    \centering
    \includegraphics[width=1\linewidth]{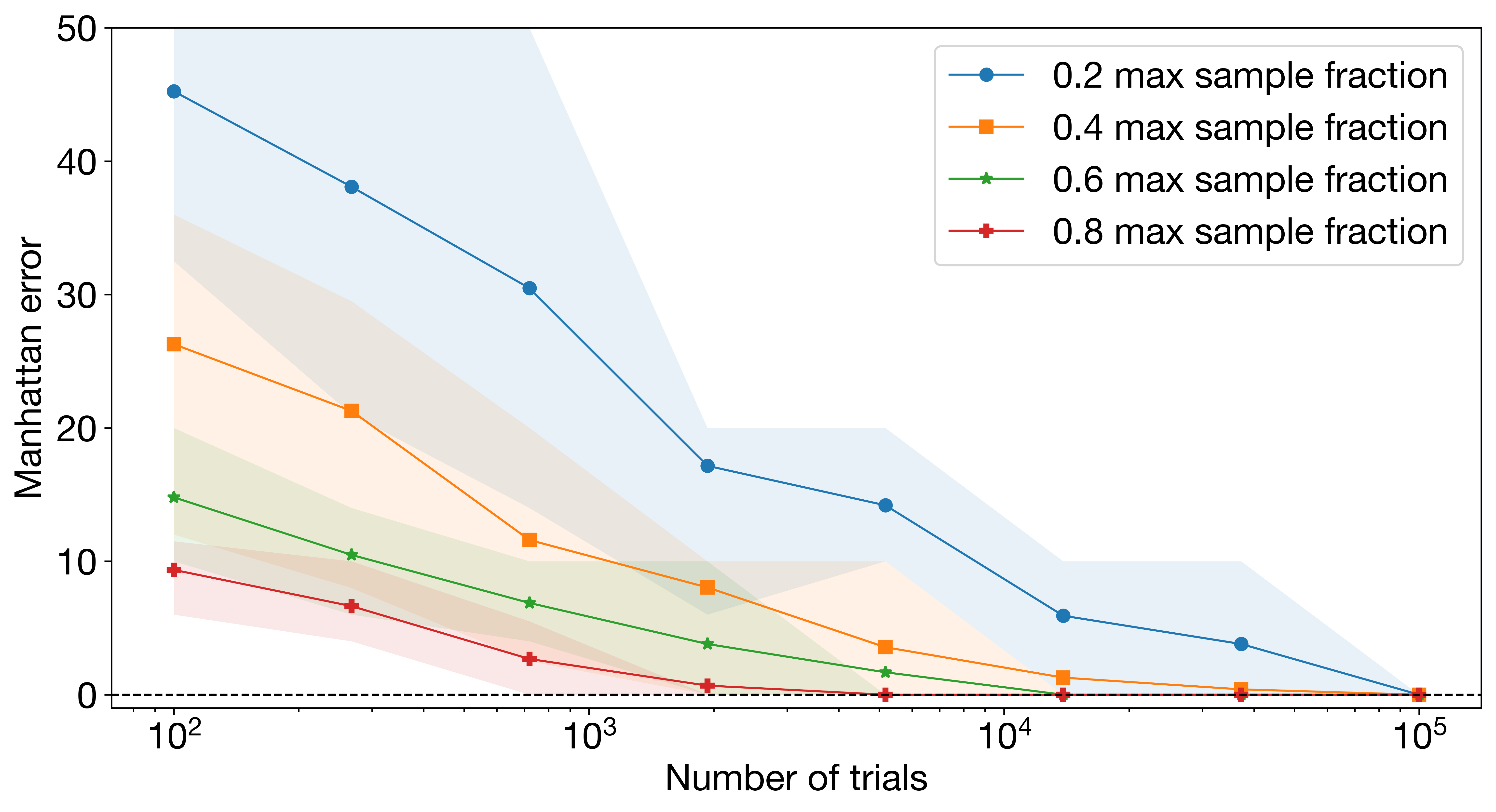}
    \caption{Maximum likelihood estimate Manhattan error for different numbers of trials at different max sample fractions. 50\% confidence interval over 50 random seeds. $K=2, N=100$ $(N_1=40, N_2=60)$}
    \label{fig:scaling_k=2}
\end{figure}

We next investigate the robustness of the maximum likelihood estimate for a wide range of numbers of observations (trials) $T$ and max sample fractions $f_{max}$. As the data is discrete, we use the Manhattan distance between the true and estimated counts to quantify the estimation error. In Figure \ref{fig:scaling_k=2}, we see that the location of the minimum NLL converges to the true value as the number of observations increases, for different levels of under-sampling.  As expected, a higher max sample fraction results in faster convergence and more accurate estimates, as quantified by lower error for the same number of samples. We extend this experiment to $K=3$ (Figure \ref{fig:scaling_k=3}, Appendix \ref{appendix:extra_figs}), and find that increasing the number of categories reduces the number of observations required to reach zero error. Here we use a confidence interval of 50\% for visual clarity, and the same figure with a 90\% confidence interval is included in Appendix \ref{appendix:extra_figs}.

\begin{figure}
    \centering
    \includegraphics[width=0.8\columnwidth]{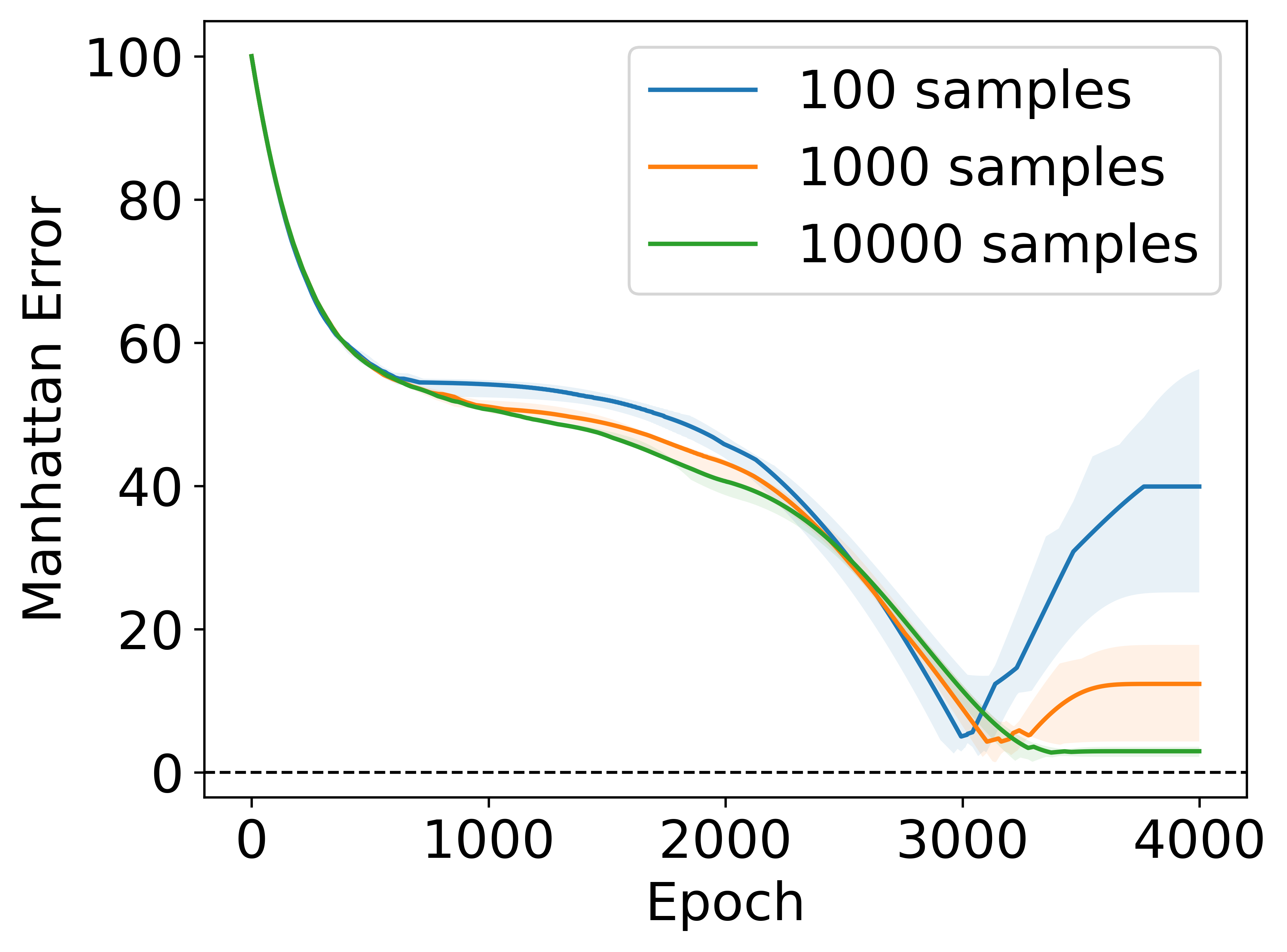}
     \caption{Maximum likelihood estimate Manhattan error per training epoch obtained with gradient descent, for different numbers of trials. The accuracy increases and the variance of the estimate decreases with increasing number of samples. The increase in error following an initial decrease occurs because we measure absolute error, and this behavior corresponds to the estimate overshooting the true value. 50\% confidence interval over 20 random seeds. $K=2, f_{max}=0.4, N=100$ $(N_1=30, N_2=70)$}
     \label{fig:sgd_k=2}
\end{figure}

Next, in Figure \ref{fig:sgd_k=2} we show that the maximum likelihood estimate can be obtained using gradient descent with the hypergeometric negative log-likelihood objective. We generate samples as before, and perform gradient descent using Adam with a learning rate of 0.1 and using a zero-initialization for the count parameters $\hat{N}_i$. We find that increasing the number of categories (Figure \ref{fig:sgd_k=3}, Appendix \ref{appendix:extra_figs}) improves the rate of convergence of the estimate and reduces the final error for the same number of samples. This provides empirical evidence that the bias of the maximum likelihood estimator decreases with increasing $K$. Additional experiments with different $f_{max}$ are included in Appendix \ref{appendix:extra_figs}, and we show representative results for a mixture of distributions in Figure \ref{fig:vae_trajectory} and Figure \ref{fig:vae_total} (Appendix \ref{appendix:extra_figs}).

\begin{table*}[t]
\caption{Representative results comparing the hypergeometric (HG), multinomial (MN) and Poisson (P) likelihood estimates on simulated datasets (additional results in Appendix \ref{appendix:extra_results}). Ground-truth distributions are uniformly under-sampled by $20-60\%$. Metrics are adjusted Rand index (ARI), median percentage error (MPE), and mean absolute error (MAE), averaged over 5 simulation random seeds.}
\label{table:results}
\vskip 0.1in
\begin{center}
\begin{scriptsize}
\begin{sc}
\begin{tabular}{l||c|c|c||c|c||c|c}
 & \multicolumn{3}{c||}{ARI} & \multicolumn{2}{c||}{MPE} & \multicolumn{2}{c}{MAE} \\
 & HG (ours) & MN & P & HG (ours) & P & HG (ours) & P \\
\hline
\makecell{3 distributions (2 unique), $10^3$ categories,\\$10^3$ obs/distribution, $10^4$ ground-truth total} & 
\makecell{$\mathbf{1.00}$\\$\pm 0.01$} & \makecell{$0.45$\\$\pm 0.01$} & \makecell{$0.45$\\$\pm 0.15$} & 
\makecell{$\mathbf{2.8}$\\$\pm 0.3\%$} & \makecell{$26.1$\\$\pm 0.1\%$} & 
\makecell{$\mathbf{416}$\\$\pm 49$} & \makecell{$3471$\\$\pm 2$} \\
\hline
\makecell{3 distributions (2 unique), $10^4$ categories,\\$10^4$ obs/distribution, $10^4$ ground-truth total} & 
\makecell{$\mathbf{1.00}$\\$\pm 0.01$} & \makecell{$0.45$\\$\pm 0.01$} & \makecell{$0.50$\\$\pm 0.01$} & 
\makecell{$\mathbf{3.9}$\\$\pm 0.9\%$} & \makecell{$23.3$\\$\pm 0.8\%$} & 
\makecell{$\mathbf{573}$\\$\pm 152$} & \makecell{$3744$\\$\pm 23$} \\
\hline
\makecell{10 distributions (9 unique), $10^3$ categories,\\$10^5$ obs/distribution, $10^5$ ground-truth total} & 
\makecell{$\mathbf{0.99}$\\$\pm 0.01$} & \makecell{$0.89$\\$\pm 0.01$} & \makecell{$0.87$\\$\pm 0.01$} & 
\makecell{$\mathbf{1.6}$\\$\pm 0.1\%$} & \makecell{$6.5$\\$\pm 0.1\%$} & 
\makecell{$\mathbf{225}$\\$\pm 6$} & \makecell{$7309$\\$\pm 43$} \\
\hline
\makecell{10 distributions, $10^1$ categories,\\$10^3$ obs/distribution, $10^3$ ground-truth total} & 
\makecell{$1.00$\\$\pm 0.01$} & \makecell{$1.00$\\$\pm 0.01$} & \makecell{$0.978$\\$\pm 0.023$} & 
\makecell{$\mathbf{2.5}$\\$\pm 0.2\%$} & \makecell{$64$\\$\pm 1\%$} & 
\makecell{$\mathbf{27}$\\$\pm 3$} & \makecell{$6.2$\\$\pm 0.1$} \\
\hline
\makecell{10 distributions, 10 categories,\\$10^5$ obs/distribution, $10^3$ ground-truth total} & 
\makecell{$1.00$\\$\pm 0.01$} & \makecell{$1.00$\\$\pm 0.01$} & \makecell{$0.78$\\$\pm 0.01$} & 
\makecell{$\mathbf{5.2}$\\$\pm 1.5\%$} & \makecell{$6.3$\\$\pm 0.1\%$} & 
\makecell{$\mathbf{52}$\\$\pm 15$} & \makecell{$63$\\$\pm 1$} \\
\hline

\end{tabular}
\end{sc}
\end{scriptsize}
\end{center}
\vskip -0.1in
\end{table*}

\subsection{Benchmarks}
Having demonstrated that the inference problem is tractable, we turn to our complete latent-variable model. To show the advantages of using the hypergeometric distribution for inference, we compare it to other common distributional assumptions.

\textit{Multinomial}: The multinomial likelihood is commonly used in recommender systems - while it cannot be used to infer absolute counts, it can used to perform unsupervised clustering and estimate relative category frequency \cite{liang_variational_2018}.

\textit{Poisson}: The Poisson distribution is a discrete distribution often used to model count data \cite{inouye2017review}, however the distribution is univariate which removes the possibility of directly capturing dependence between features. To account for sparsity caused by under-sampling, a zero-inflation component is sometimes added to account for excess zeros \cite{lambert1992zero}, but that is unnecessary here as we know our simulated datasets are not zero-inflated.

To directly compare the impact of likelihood choice, we use the exact same VAE architecture and training procedure and swap out the likelihood models.

\subsubsection{Count Estimation Experiment}
Since we are proposing a generative model for count data, we first show that our VAE with hypergeometric likelihood is able to better estimate the data than using alternate likelihoods. We simulate datasets with K = 100, 1000, or 10,000 categories, each consisting of a mixture of either 3 or 10 discrete distributions (simulation details in Appendix \ref{appendix:simulation}). We train the base model with each likelihood until convergence and then quantify the estimation error as the discrepancy between the ground truth distribution and the model's estimate, using both mean absolute error (MAE) and median percentage error (MPE) metrics.

Representative results are shown in Table \ref{table:results}, with further results in Appendix \ref{appendix:extra_results}. Our method clearly outperforms the Poisson likelihood in its ability to correctly estimate the under-sampled ground truth distribution, and produces estimates with significantly lower absolute and percent error.

\subsubsection{Clustering Experiment}
We next show that the hypergeometric likelihood can correctly infer the latent structure in cases where other likelihood functions fail. A major benefit of using a latent variable model is that we can make use of the learned latent representation $\vz$ to perform deep unsupervised clustering \cite{lim2020deep}. Both the multinomial and Poisson likelihoods have been used to learn useful latent representations of count data \cite{bouguila2010count}. The assumptions of the hypergeometric distribution, which considers both absolute and relative counts, can be thought of as a superset of the assumptions of the multinomial distribution, which only considers relative counts. For example, this difference would become important in datasets composed of a mixture of distributions with similar count fractions but different total counts.

For certain simulated datasets from the preceding experiment, we assign two of the count distributions to have a different number of total elements but similar probability distributions over categories. As a practical example, these could represent two similar ecosystems with comparable fractions of animal species, but different total populations due to differences in climate. We train our base model with different likelihoods on these datasets in an unsupervised fashion. We cluster the final latent representation of the samples using k-means (with the true number of distributions), and compare the cluster labels to the ground-truth. We use the adjusted Rand index (ARI) as the clustering metric \citep{santos2009use}, taking the best result over ten random centroid initializations.

Representative results for this experiment are again shown in Table \ref{table:results}. Our method outperforms all other likelihoods in its ability to learn a latent space that is able to differentiate between underlying distributions. In particular, it is the only examined method able to distinguish between count distributions with similar probability distributions over features. These results are consistent across a wide variety of simulated datasets (Appendix \ref{appendix:extra_results}).

\subsection{Application to Reading Passage Complexity}
\label{sec:complexity}

In the field of NLP, it is valuable to be able to measure the linguistic complexity of a text. For example, quantifying the readability of text passages is essential in the education context, as it allows texts to be correctly matched to students at the appropriate level of reading skill \cite{sarti2021looks}. We hypothesize that any short text passage can be thought of as a bag-of-words (BoW) that is sampled from a larger latent BoW, and that the size and complexity of this underlying BoW provides a better measure of the readability of a passage than the passage itself. That is, a text passage that is generated from a richer underlying vocabulary will demand a higher level of reading comprehension from its reader.

We test this hypothesis using the CommonLit Ease of Readability (CLEAR) Corpus \citep{crossley2023large}, an open-source dataset consisting of almost 5000 text excerpts sourced from Grade 3-12 reading curricula. Numerous metrics have been proposed to estimate the complexity of reading passages, and each excerpt has been annotated with a number of well-established readability indices, including those can be automatically calculated by ARTE, the Automatic Readability Tools for English \citep{choi2022advances}. Additionally, each passage has a Bradley-Terry reading ease score \cite{bradley1952rank} derived from the qualitative assessment by teachers of the relative reading difficulty between pairs of passages.

This scenario presents an ideal application of our proposed hypergeometric method, which can capture both the dependence between the choice of words (the selection of one word comes at the cost of its synonyms, and to the benefit of words from the same topic) and the vocabulary size as measured in token counts. Additionally, since there is no clear classification of the texts, a continuous latent variable model is well suited. We create a bag-of-words representation by tokenizing the 4724 text excerpts in the CLEAR Corpus, producing a count matrix over the resulting tokens (Appendix \ref{appendix:complexity} for details). We then train our unsupervised model on this BoW representation of the excerpts, which enables us to infer a latent BoW that underlies each passage. We use this estimate as a proxy for passage readability using both the \textit{total} estimated number of tokens $t_{total}$ and the number of \textit{unique} tokens $t_{unique}$ (tokens with non-zero count) in each passage's latent BoW. We then calculate the Pearson correlation across the 4724 passages between each of our proposed metrics ($t_{total}$ and $t_{unique}$) and the readability indices provided in the CLEAR dataset, in order to assess the association strength.

\begin{figure}[t]
     \centering
     \begin{subfigure}{\columnwidth}
         \centering
         \includegraphics[width=0.9\columnwidth]{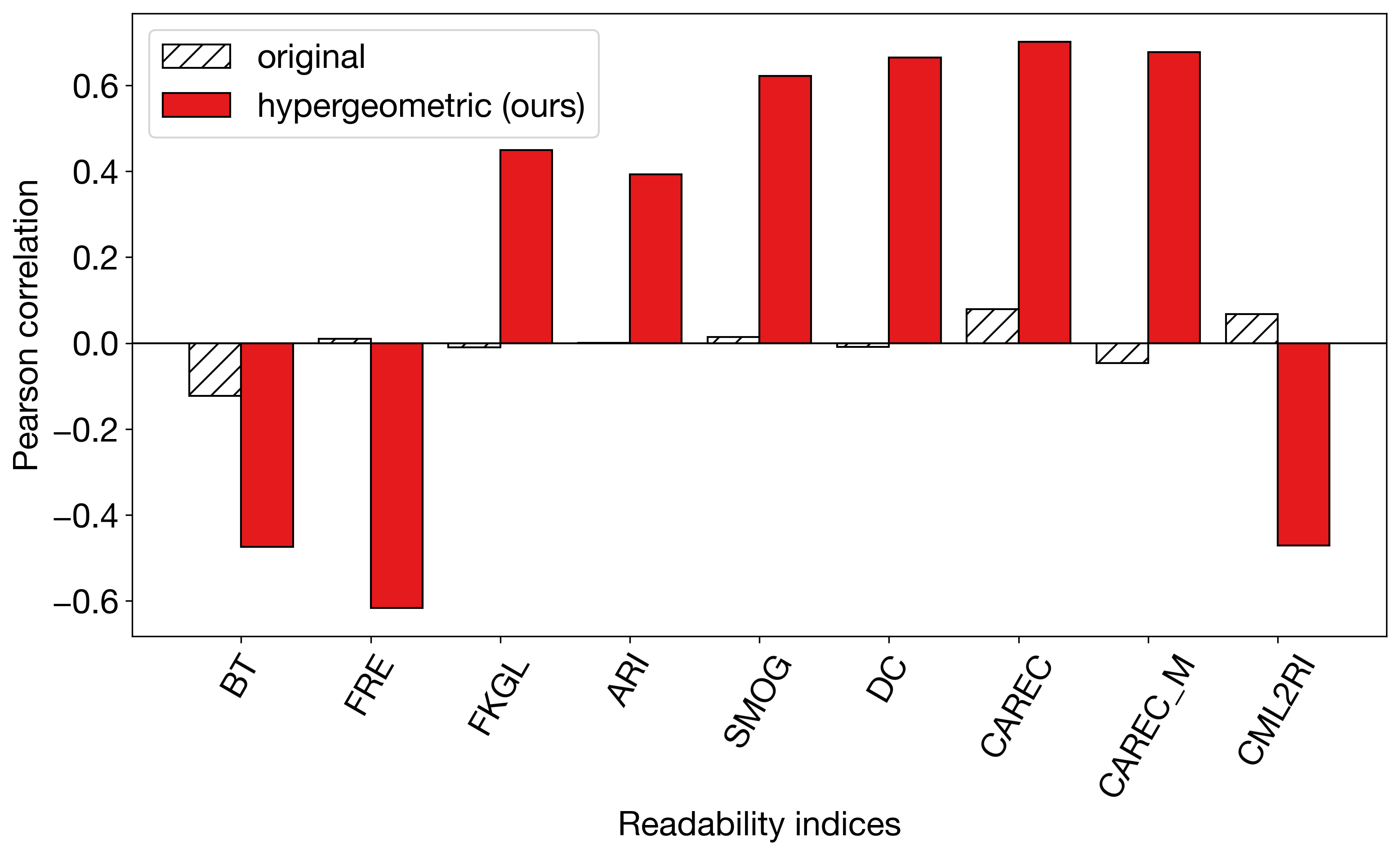}
         \caption{Total number of tokens ($t_{total}$)}
         \label{fig:complexity_total}
     \end{subfigure}
     \vfill
     \begin{subfigure}{\columnwidth}
         \centering
         \includegraphics[width=0.9\columnwidth]{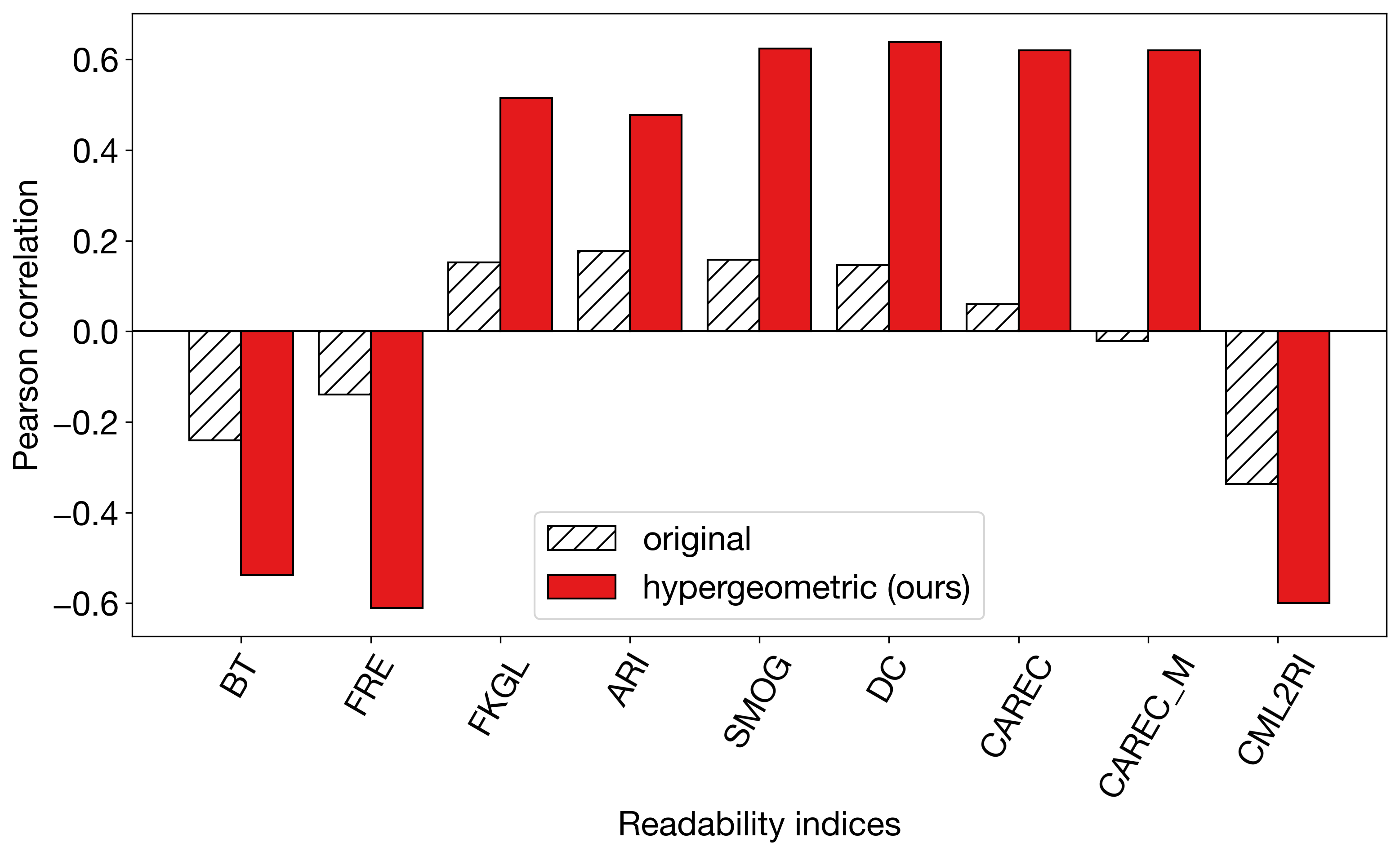}
         \caption{Unique number of tokens ($t_{unique}$)}
         \label{fig:complexity_unique}
     \end{subfigure}
     \caption{Comparison of association strength between reading passage complexity metrics and total/unique tokens in the original/latent bag-of-words. Readability indices are BT easiness (BT), Flesch-Reading-Ease (FRE), Flesch-Kincaid-Grade-Level (FKGL), Automated Readability Index (ARI), SMOG Readability (SMOG), New Dale-Chall Readability Formula (DCR), CAREC, CAREC-M, and CML2RI.}
     \label{fig:complexity_correlations}
\end{figure}

Figure \ref{fig:complexity_correlations} shows the correlations between established readability indices and our proposed complexity metrics. In Figure \ref{fig:complexity_total}, we see that initially the there is a near-zero association between the total number of tokens in the BoW representation of the passages and the readability indices before applying our model. This demonstrates that more complex excerpts do not have a larger token count in their raw form, and hence this is not a good naive metric for readability. However, our model's estimate for the size of the latent BoW exhibits a strong association with all of the reading complexity metrics. Therefore, our model successfully infers that a larger latent BoW underlies the more complex reading passages. Note that for metrics BT, FRE and CML2RI, a higher value indicates a less complex passage, hence the negative correlations.

Similarly, Figure \ref{fig:complexity_unique} shows that the vocabulary size of the latent BoW (the number of unique non-zero tokens for a passage) inferred by our model also aligns with the complexity of the passage much better than if we used the original text's vocabulary. These results demonstrate that our method is able to infer a latent vocabulary underlying each reading passage that convincingly reflects the independently-assessed readability of the text.

\subsection{Application to Single-Cell Genomics}
\label{sec:sc}
The detailed measurement of the contents of individual cells promises to vastly improve our understanding of fundamental biology. In recent years, the advancement of high-throughput techniques in the field of single-cell genomics has enabled the collection of large numbers of gene transcripts from individual cells, resulting in vast count matrices. Each cell has a finite population of transcripts that can be captured - on the order of one hundred thousand to a million - and a transcript can only be captured once prior to sequencing \cite{de2019chromium}. It is accepted that the main source of technical noise in this data is due to under-sampling \cite{kuo_quantification_2022}, leading to the well-known phenomenon of dropout (inflated occurrence of zeros in the final count matrix). This technical noise hinders scientists' ability to draw meaningful scientific conclusions from these experiments.

The hypergeometric distribution is well-suited to modelling this capture process, as the number of captured gene transcripts in an experiment is significant relative to the total population size, and capture occurs without replacement, leading to dependence between gene feature counts. The high-dimensional distribution of gene transcript counts can be effectively represented by latent variable models \cite{lopez_deep_2018,zhao2021learning}, as individual genes are often members of co-expression networks which result in highly correlated counts. We therefore use our method, with genes as categories, in order to infer the true gene transcripts counts in each cell from the sparse, under-sampled count data.

Because there is typically no way of knowing the true number of transcripts of each gene in a given cell, we focus on what is known as a spike-in experiment, which does provide a ground-truth. In the experiment we consider \citep{ziegenhain_molecular_2022}, a known concentration of a solution of synthetic RNA is placed in small wells, and human cells are individually placed in a subset of these wells. After the transcripts present in the wells have been captured and sequenced, we obtain experimental counts corresponding to both the human RNA (for which the true amount is unknown) and the synthetic RNA (true amount is known). The wells that did not have cells in them should have an equal amount of the synthetic RNA across measurements, so we can therefore use the measured counts of synthetic RNA in both empty and cell-containing wells as a ground-truth reference to evaluate our estimated human gene transcript counts.

\textit{Dataset}: The SPIKE dataset consists of counts for 43k genes (categories) across 1126 observations (cells). Of these, 181 observations contain a mixture of synthetic RNA and human kidney cells (labelled ``HEK293T"), and the remainder 945 contain only the synthetic RNA (labelled ``empty").

\begin{figure}[h]
    \centering
    \includegraphics[width=0.85\columnwidth]{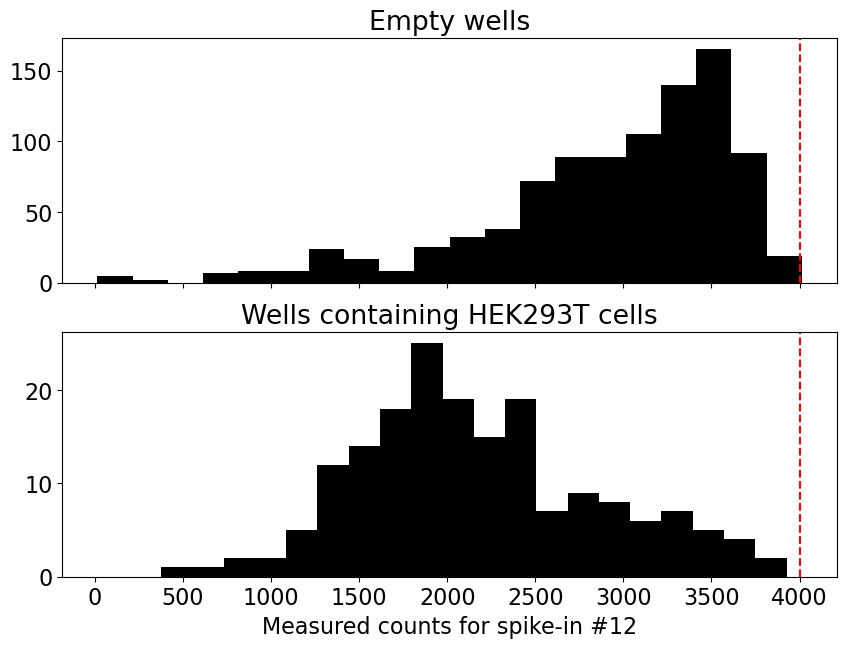}
    \caption{Measured counts for synthetic spike-in RNA \#12 with and without human cells present. Red dashed line is ground-truth count (ideally all measurements would be equal).}
    \label{fig:measured_sp12}
\end{figure}

Figure \ref{fig:measured_sp12} compares the distribution of measured counts in this dataset for one specific synthetic RNA (\#12) for the empty and HEK293T measurements. This difference between these distributions exemplifies the stochastic under-sampling that occurs, as the true counts should be identical across all observations - approximately 4000 for this particular spike-in RNA. It is also clear that the presence of the additional human RNA in the HEK293T observations significantly lowers the amount of synthetic RNA that is captured, which supports the assumption that there is dependence between counts, i.e. a transcript is captured at the cost of another.

\begin{figure}[h!]
     \centering
     \begin{subfigure}{\columnwidth}
         \centering
         \includegraphics[width=0.85\columnwidth]{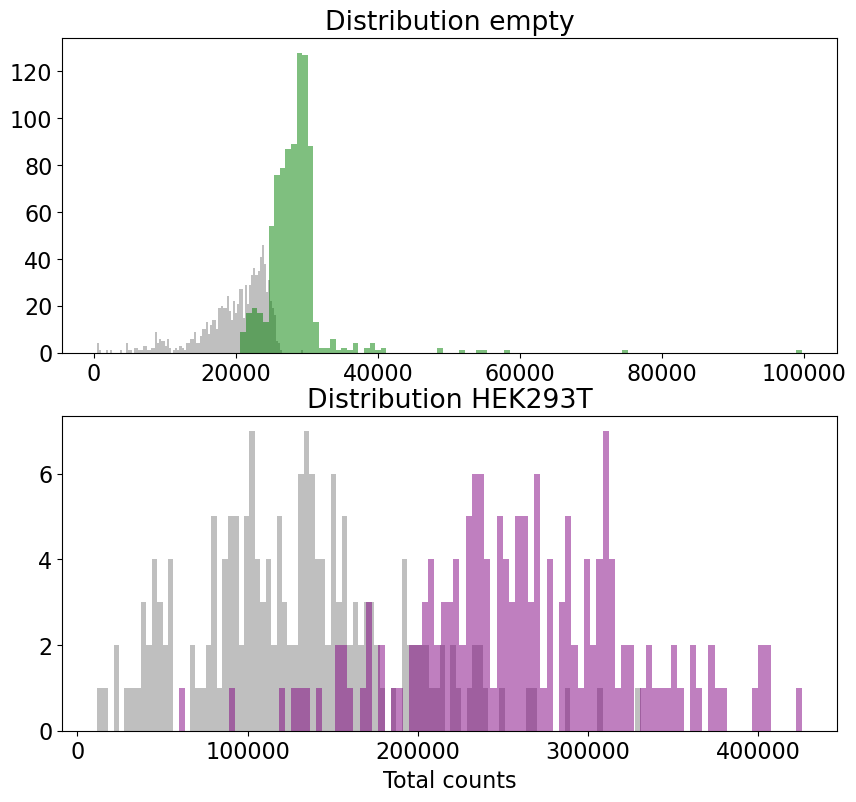}
         \caption{Total counts}
         \label{fig:spike_totals}
     \end{subfigure}
     \vfill
     \begin{subfigure}{\columnwidth}
         \centering
         \includegraphics[width=0.85\columnwidth]{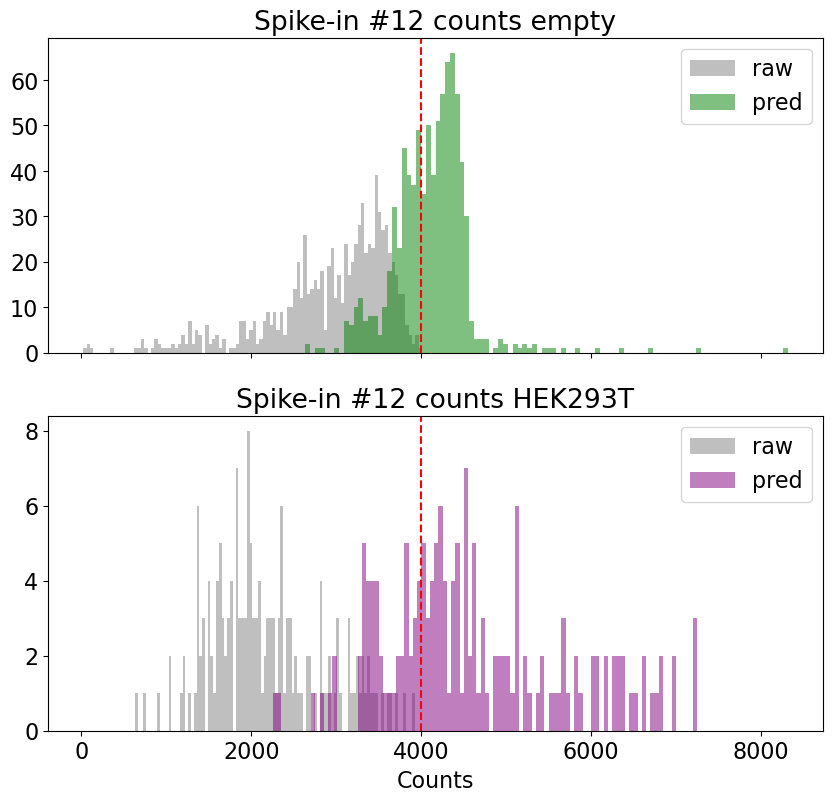}
         \caption{Spike-in \#12 counts}
         \label{fig:spike12_totals}
     \end{subfigure}
     \caption{Histograms of the number of transcript counts when only synthetic RNA is present (top) and when both synthetic RNA and human RNA (from kidney cells) are present (bottom). The original (measured) distribution is in grey, and the red dashed line shows the ground-truth amount of synthetic RNA \#12.}
     \label{fig:spike_estimates}
\end{figure}

We train our hypergeometric model on the SPIKE dataset to infer the ground-truth count distribution. We use the top $K=10,000$ genes (categories) by mean transcript count, for computational efficiency and because many genes are not expressed in this celltype. Model and training hyperparameters are given in Appendix \ref{appendix:hyperparams}. Summary results of the final estimated count matrix are shown in Figure \ref{fig:spike_estimates}.

Figure \ref{fig:spike_totals} shows the inferred total number of counts per measurement well (the sum of the estimated count matrix rows). We can see that the variance of the total number of spike-in RNA, which is known to be constant, is significantly reduced in our estimate. The experiment's authors estimated the total amount of synthetic RNA per observation as approximately 30,000, which aligns with our model estimate. The unknown total amount of RNA in the human kidney cells is estimated by our model to have a median of 260,000, in line with a previous estimate of 200,000 \cite{shapiro2013single}. Note that the number of observations available in this dataset is very small compared to typical single-cell datasets (without ground-truth), and this estimate is expected to be even more accurate and have a lower variance with a larger number of observed cells. Figure \ref{fig:spike12_totals} shows the estimated counts specifically for spike-in \#12, whose ground-truth value is approximately 4000. We can see that although the initial distribution of counts for this synthetic RNA has a significantly different mean in the empty and cell-containing observations, our model produces an estimate near the true value, and brings both distributions into alignment (this is also true in the other spike-in RNA). These results show that we are able to recover the known ground-truth distribution of synthetic RNA that has been corrupted by the stochastic capture process.

\section{Conclusion}
We propose a method for estimating the unknown population sizes for a mixture of discrete distributions of elements using the hypergeometric likelihood. Through empirical data simulation we show for the first time that inference of the true population size is tractable over a range of maximum sample fractions, and that our method outperforms other common distributional assumptions used for count data. We address the real-world problems of estimating reading passage complexity and reversing gene transcript count sparsity, and show that our method is able to recover the true number of transcripts in a cell. Due to the prevalence of finitely sampled discrete populations in biology and other matrix completion settings, we expect this method can be successfully used in many other application domains.

\section*{Impact Statement}
This paper presents work whose goal is to advance the field of machine learning, in particular through its applications to genomics experimental data.

\bibliography{icml2024}
\bibliographystyle{icml2024}

\newpage
\appendix
\onecolumn

\section{Data Simulation}
\label{appendix:simulation}

\begin{algorithm*}[h]
   \caption{Dataset simulation}
   \label{alg:example}
\begin{algorithmic}
   \STATE {\bfseries Input:} \# distributions $M$, \# categories $K$, \# trials (observations/distribution) $T$, true total counts per distribution $N$, sample depths $f_{min}$,$f_{max}$
   \FOR{each distribution $m \in M$}
   \STATE Draw distribution $p$ over $K$ categories ($p \sim \text{Dirichlet}$)
   \STATE Ground-truth count distribution $P = p*N$ (round to nearest integer)
   \FOR{$T$ observations}
   \STATE Draw sample depth $n$ from Uniform($f_{min} * N$, $f_{max} * N$)
   \STATE Draw $n$ samples from ground-truth count distribution $P$ without replacement
   \ENDFOR
   \ENDFOR
\end{algorithmic}
\end{algorithm*}

\section{Training Hyperparameters}
\label{appendix:hyperparams}

\begin{table}[h]
\centering
\caption{Model hyperparameters}
\label{sample-table}
\begin{tabular}{l|lll}
\bf Parameter & \bf Simulated & \bf CLEAR & \bf SPIKE
\\ \hline \\
Encoder layers & 128, 128 & 128, 128 & 128, 128 \\
Decoder layers & 128, 128 & 128, 128 & 256, 256 \\
Latent space dimension & 10 & 10 & 16 \\
Learning rate & 0.01 & 0.001 & 0.01 \\
Batch size & 100 & 100 & 563 \\
Violation penalty (min/max) & 1 & 1 & 1/100 \\

\end{tabular}
\end{table}

\section{CLEAR Reading Passage Complexity Implementation Details}
\label{appendix:complexity}

We download v6.01 of the CLEAR dataset, which contain 4724 excerpts from books and other documents. We tokenize all passages using the scikit-learn function CountVectorizer(strip\_accents='unicode'), resulting in a sparse array $X$ of counts for 34,015 tokens across the passages corpus. We use this as input to our model, training until convergence, and obtain the in-sample inferred vocabulary counts $\hat{X}$. We calculate the total number of tokens $t_{total}$ as the row-sum of $X$ or $\hat{X}$, and the unique tokens $t_{unique}$ as the row-sum of the binarized arrays.

\newpage
\section{Additional Results}
\label{appendix:extra_results}

\begin{table*}[h]
\caption{Representative results comparing the hypergeometric (HG), multinomial (MN) and Poisson (P) likelihood estimates on simulated datasets (additional results in Appendix \ref{appendix:extra_results}). Ground-truth distributions are uniformly under-sampled by $20-60\%$. Metrics are adjusted Rand index (ARI), median percentage error (MPE), and mean absolute error (MAE), averaged over 5 simulation random seeds.}
\label{table:results}
\vskip 0.1in
\begin{center}
\begin{scriptsize}
\begin{sc}
\begin{tabular}{l||c|c|c||c|c||c|c}
 & \multicolumn{3}{c||}{ARI} & \multicolumn{2}{c||}{MPE} & \multicolumn{2}{c}{MAE} \\
 & HG (ours) & MN & P & HG (ours) & P & HG (ours) & P \\
\hline
\makecell{3 distributions (2 unique), $10^3$ categories,\\$10^3$ obs/distribution, $10^4$ ground-truth total} & 
\makecell{$1.00$\\$\pm 0.01$} & \makecell{$0.45$\\$\pm 0.01$} & \makecell{$0.45$\\$\pm 0.15$} & 
\makecell{$2.8$\\$\pm 0.3\%$} & \makecell{$26.1$\\$\pm 0.1\%$} & 
\makecell{$416$\\$\pm 49$} & \makecell{$3471$\\$\pm 2$} \\
\hline
\makecell{3 distributions (2 unique), $10^4$ categories,\\$10^4$ obs/distribution, $10^4$ ground-truth total} & 
\makecell{$1.00$\\$\pm 0.01$} & \makecell{$0.45$\\$\pm 0.01$} & \makecell{$0.50$\\$\pm 0.01$} & 
\makecell{$3.9$\\$\pm 0.9\%$} & \makecell{$23.3$\\$\pm 0.8\%$} & 
\makecell{$573$\\$\pm 152$} & \makecell{$3744$\\$\pm 23$} \\
\hline
\makecell{10 distributions (9 unique), $10^3$ categories,\\$10^5$ obs/distribution, $10^5$ ground-truth total} & 
\makecell{$0.99$\\$\pm 0.01$} & \makecell{$0.89$\\$\pm 0.01$} & \makecell{$0.87$\\$\pm 0.01$} & 
\makecell{$1.6$\\$\pm 0.1\%$} & \makecell{$6.5$\\$\pm 0.1\%$} & 
\makecell{$225$\\$\pm 6$} & \makecell{$7309$\\$\pm 43$} \\
\hline
\makecell{10 distributions, $10^1$ categories,\\$10^3$ obs/distribution, $10^3$ ground-truth total} & 
\makecell{$1.00$\\$\pm 0.01$} & \makecell{$1.00$\\$\pm 0.01$} & \makecell{$0.978$\\$\pm 0.023$} & 
\makecell{$2.5$\\$\pm 0.2\%$} & \makecell{$64$\\$\pm 1\%$} & 
\makecell{$27$\\$\pm 3$} & \makecell{$6.2$\\$\pm 0.1$} \\
\hline
\makecell{100 distributions, $10^3$ categories,\\$10^4$ obs/distribution, $10^4$ ground-truth total} & 
\makecell{$0.97$\\$\pm 0.01$} & \makecell{$0.10$\\$\pm 0.04$} & \makecell{$0.98$\\$\pm 0.01$} & 
\makecell{$0.8$\\$\pm 0.1\%$} & \makecell{$0.7$\\$\pm 0.1\%$} & 
\makecell{$99$\\$\pm 11$} & \makecell{$65$\\$\pm 1$} \\
\hline
\makecell{3 distributions (2 unique), $10^3$ categories,\\$10^3$ obs/distribution, $10^4$ ground-truth total} & 
\makecell{$0.89$\\$\pm 0.18$} & \makecell{$0.50$\\$\pm 0.01$} & \makecell{$0.50$\\$\pm 0.01$} & 
\makecell{$6.6$\\$\pm 1.2\%$} & \makecell{$23.0$\\$\pm 0.2\%$} & 
\makecell{$1438$\\$\pm 101$} & \makecell{$3772$\\$\pm 17$} \\
\hline
\makecell{10 distributions, $10^3$ categories,\\$10^3$ obs/distribution, $10^4$ ground-truth total} & 
\makecell{$1.00$\\$\pm 0.01$} & \makecell{$1.00$\\$\pm 0.01$} & \makecell{$0.72$\\$\pm 0.11$} & 
\makecell{$4.8$\\$\pm 1.2\%$} & \makecell{$10.3$\\$\pm 0.1\%$} & 
\makecell{$485$\\$\pm 119$} & \makecell{$943$\\$\pm 13$} \\
\hline
\makecell{10 distributions, $10^3$ categories,\\$10^3$ obs/distribution, $10^5$ ground-truth total} & 
\makecell{$1.00$\\$\pm 0.01$} & \makecell{$1.00$\\$\pm 0.01$} & \makecell{$0.51$\\$\pm 0.11$} & 
\makecell{$9.0$\\$\pm 5.5\%$} & \makecell{$10.3$\\$\pm 0.1\%$} & 
\makecell{$8767$\\$\pm 5331$} & \makecell{$9538$\\$\pm 86$} \\
\hline
\makecell{10 distributions, 10 categories,\\$10^5$ obs/distribution, $10^3$ ground-truth total} & 
\makecell{$1.00$\\$\pm 0.01$} & \makecell{$1.00$\\$\pm 0.01$} & \makecell{$0.78$\\$\pm 0.01$} & 
\makecell{$5.2$\\$\pm 1.5\%$} & \makecell{$6.3$\\$\pm 0.1\%$} & 
\makecell{$52$\\$\pm 15$} & \makecell{$63$\\$\pm 1$} \\
\hline
\makecell{10 distributions, 100 categories,\\$10^4$ obs/distribution, $10^4$ ground-truth total} & 
\makecell{$1.00$\\$\pm 0.01$} & \makecell{$1.00$\\$\pm 0.01$} & \makecell{$0.89$\\$\pm 0.01$} & 
\makecell{$4.8$\\$\pm 0.1\%$} & \makecell{$6.3$\\$\pm 0.1\%$} & 
\makecell{$481$\\$\pm 12$} & \makecell{$633$\\$\pm 1$} \\
\hline
\makecell{10 distributions (9 unique), 10 categories,\\$10^4$ obs/distribution, $10^3$ ground-truth total} & 
\makecell{$0.89$\\$\pm 0.01$} & \makecell{$0.89$\\$\pm 0.01$} & \makecell{$0.89$\\$\pm 0.01$} & 
\makecell{$3.5$\\$\pm 0.1\%$} & \makecell{$6.5$\\$\pm 0.1\%$} & 
\makecell{$48$\\$\pm 12$} & \makecell{$73$\\$\pm 1$} \\
\hline

\end{tabular}
\end{sc}
\end{scriptsize}
\end{center}
\vskip -0.1in
\end{table*}

\section{Additional Figures}
\label{appendix:extra_figs}

\begin{figure}[h]
    \centering
    \includegraphics[width=0.5\linewidth]{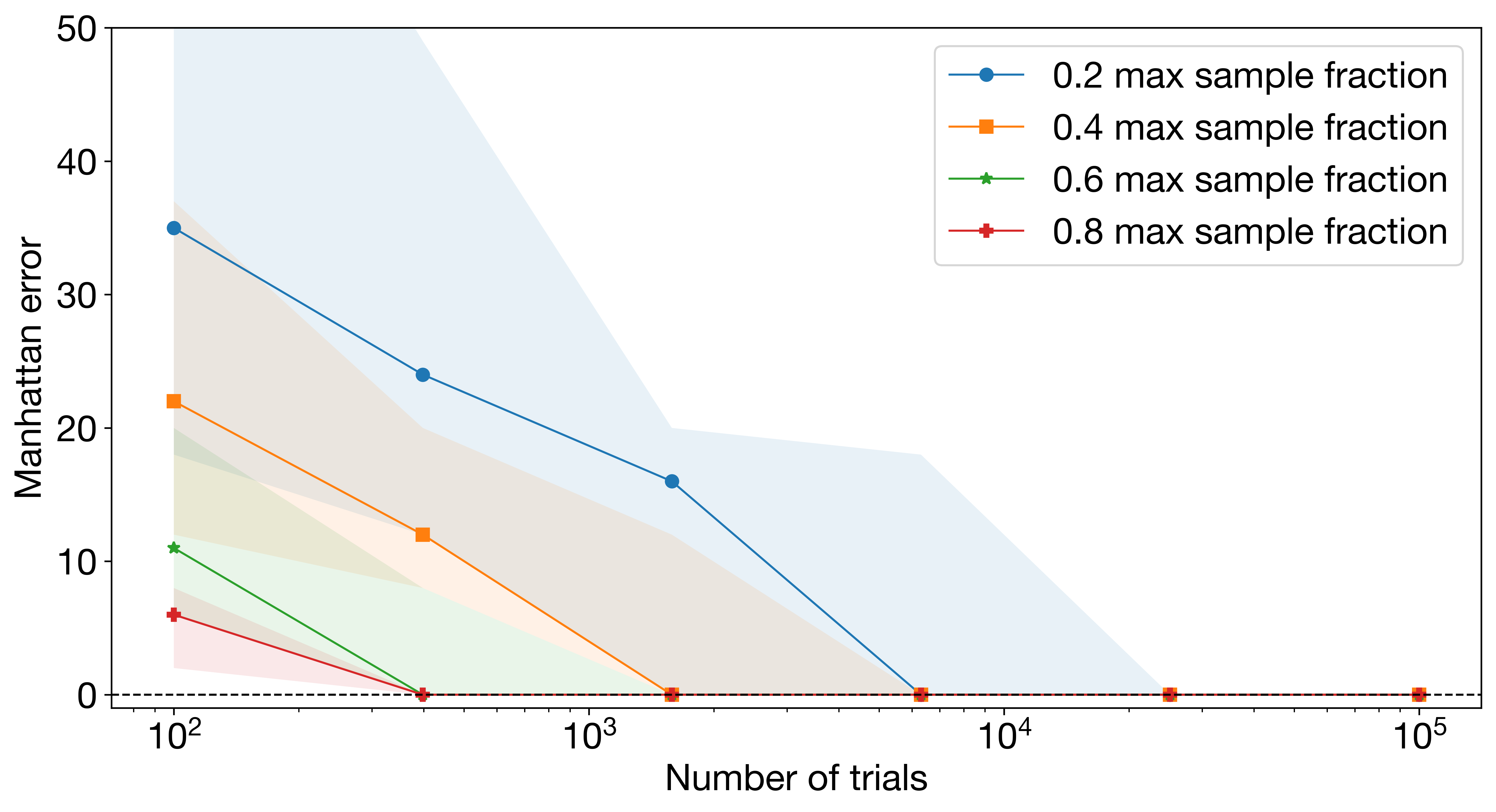}
         \caption{Maximum likelihood estimate Manhattan error for different numbers of observations at different max sample fractions ($N=100$). 50\% confidence interval over 50 random seeds. $\textbf{K=3}, N=100$ $(N_1=50, N_2=30, N_3=20)$}
         \label{fig:scaling_k=3}
\end{figure}

\begin{figure}[h]
    \centering
    \includegraphics[width=0.5\columnwidth]{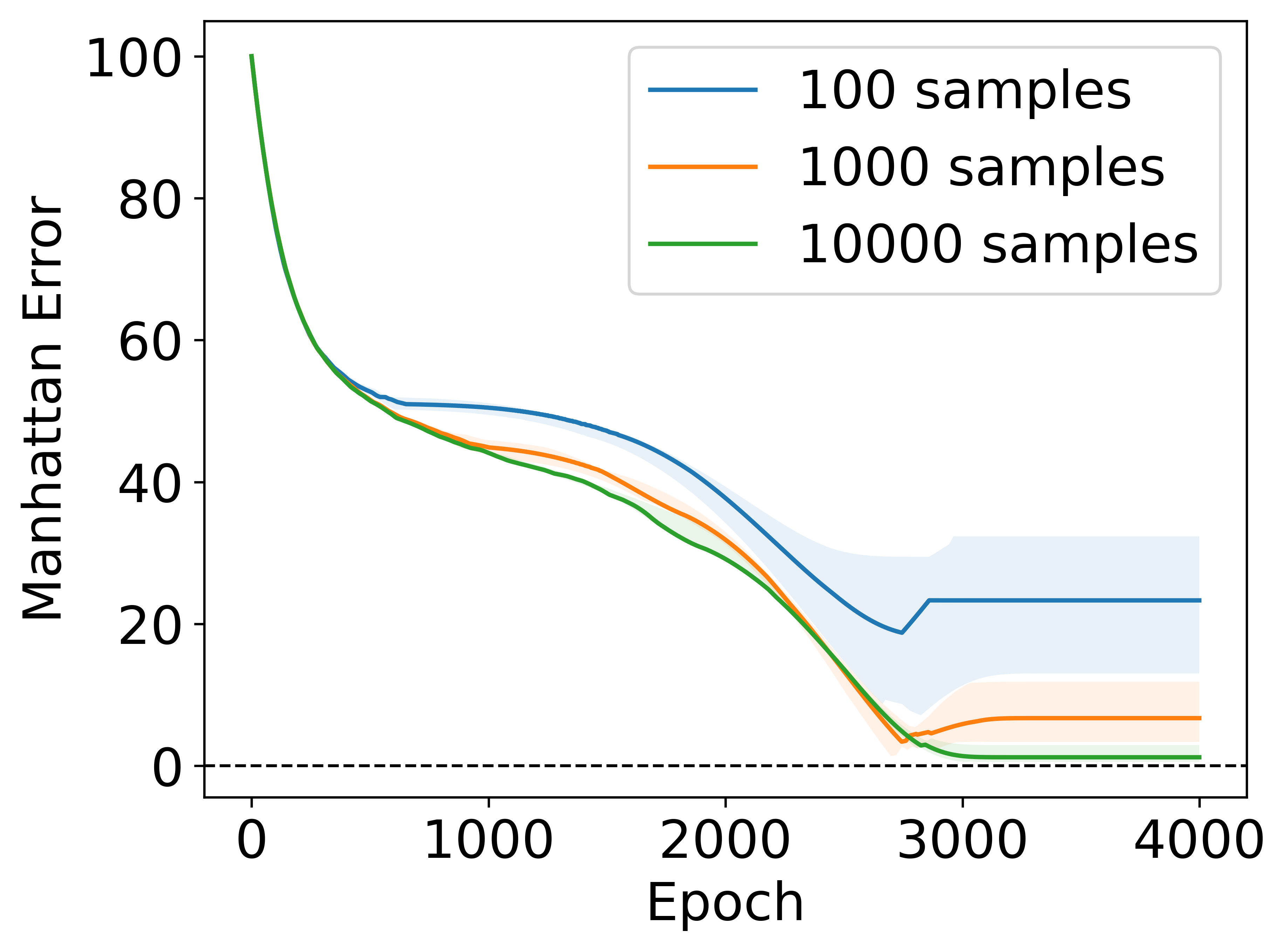}
     \caption{Maximum likelihood estimate Manhattan error per training epoch for different numbers of trials ($N=100$, max sample fraction 0.4). The apparent increase in error following an initial decrease occurs because we measure absolute error, and this behavior corresponds to the estimate approaching and overshooting the true value. 50\% confidence interval over 20 random seeds. $\textbf{K=3}, N=100$ $(N_1=50, N_2=30, N_3=20)$}
     \label{fig:sgd_k=3}
\end{figure}

\begin{figure}[h]
    \centering
    \includegraphics[width=0.5\columnwidth]{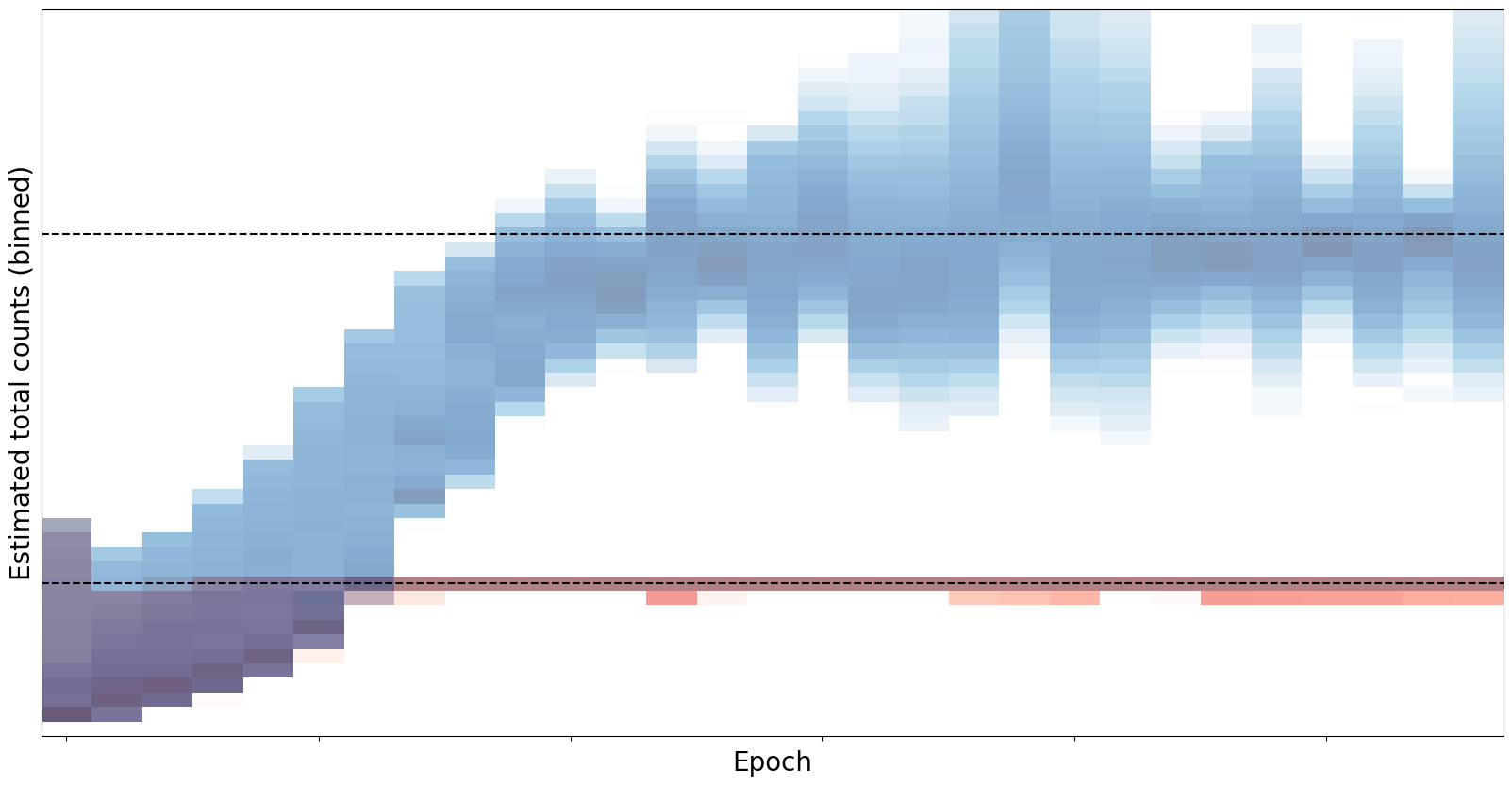}
    \caption{Vertical binning of the estimated total counts per observation vs training epochs. Black lines are true total counts for the two distributions. The final estimate recovers the ground-truth population sizes for both distributions. We emphasize that despite the model not known the true number of distributions, and therefore not having access to the labels of the observations, it is correctly able to learn a latent space that perfectly separates the two sets of samples. Note that both distributions are sampled to the same $n_{max}$, so this distinction is not due simply to differences in total observed counts for each distribution}
    \label{fig:vae_trajectory}
\end{figure}

\begin{figure}[h]
     \centering
     \begin{subfigure}{\columnwidth}
         \centering
         \includegraphics[width=0.5\columnwidth]{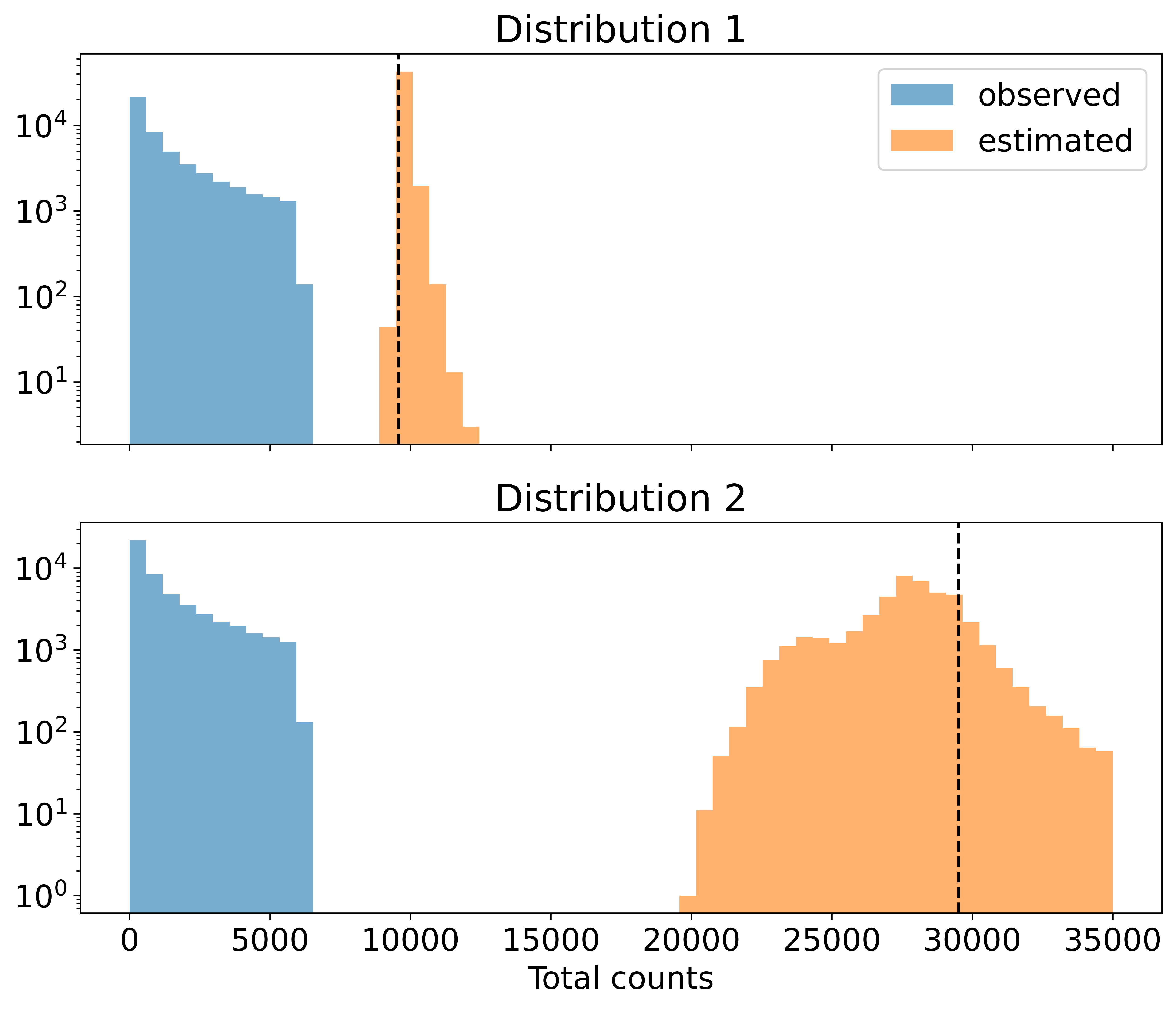}
         \caption{Total counts per observation.}
         \label{fig:vae_total_counts}
     \end{subfigure}
     \vfill
     \begin{subfigure}{\columnwidth}
         \centering
         \includegraphics[width=0.5\columnwidth]{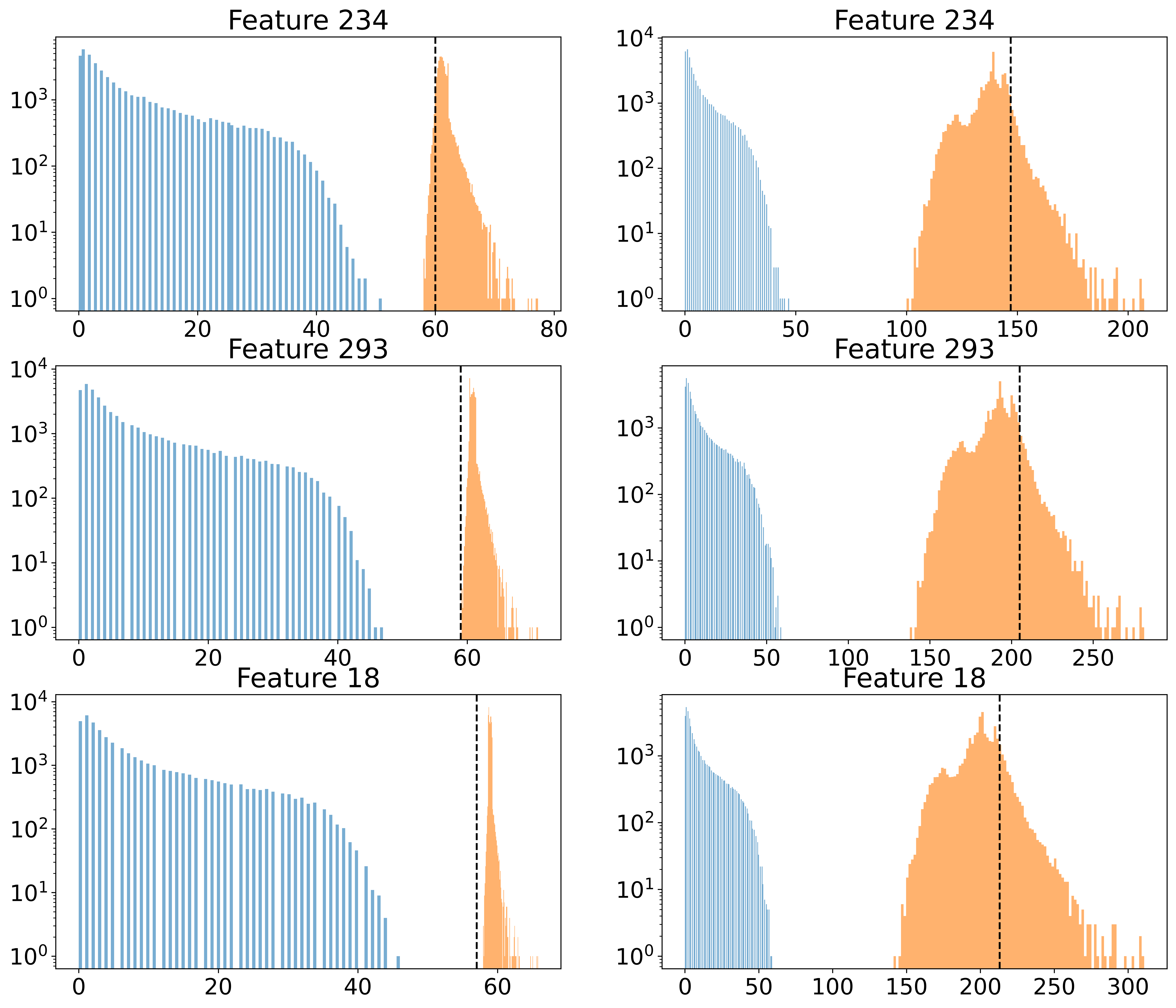}
         \caption{Counts for a subset of features (categories) for distribution 1 (left column) and 2 (right column).}
         \label{fig:vae_feature_counts}
     \end{subfigure}
     \caption{Histograms comparing the observed (blue) and estimated (orange) counts to the ground-truth underlying distribution. Our model estimate shifts the count distributions away from zero and close to ground-truth value (dashed line). For the two ground-truth distributions with different $N$, the estimates are shifted away from the observed distribution and approach the true values. We see that the variance of the estimate is higher when the under-sampling is more drastic (20\% vs 60\%). Figure \ref{fig:vae_feature_counts} shows the original and estimated count distributions for the top three categories by mean ground-truth count, again showing that the estimates approach the ground-truth values.}
     \label{fig:vae_total}
\end{figure}

\begin{figure}[h]
     \centering
     \begin{subfigure}[b]{0.49\textwidth}
         \centering
         \includegraphics[width=\textwidth]{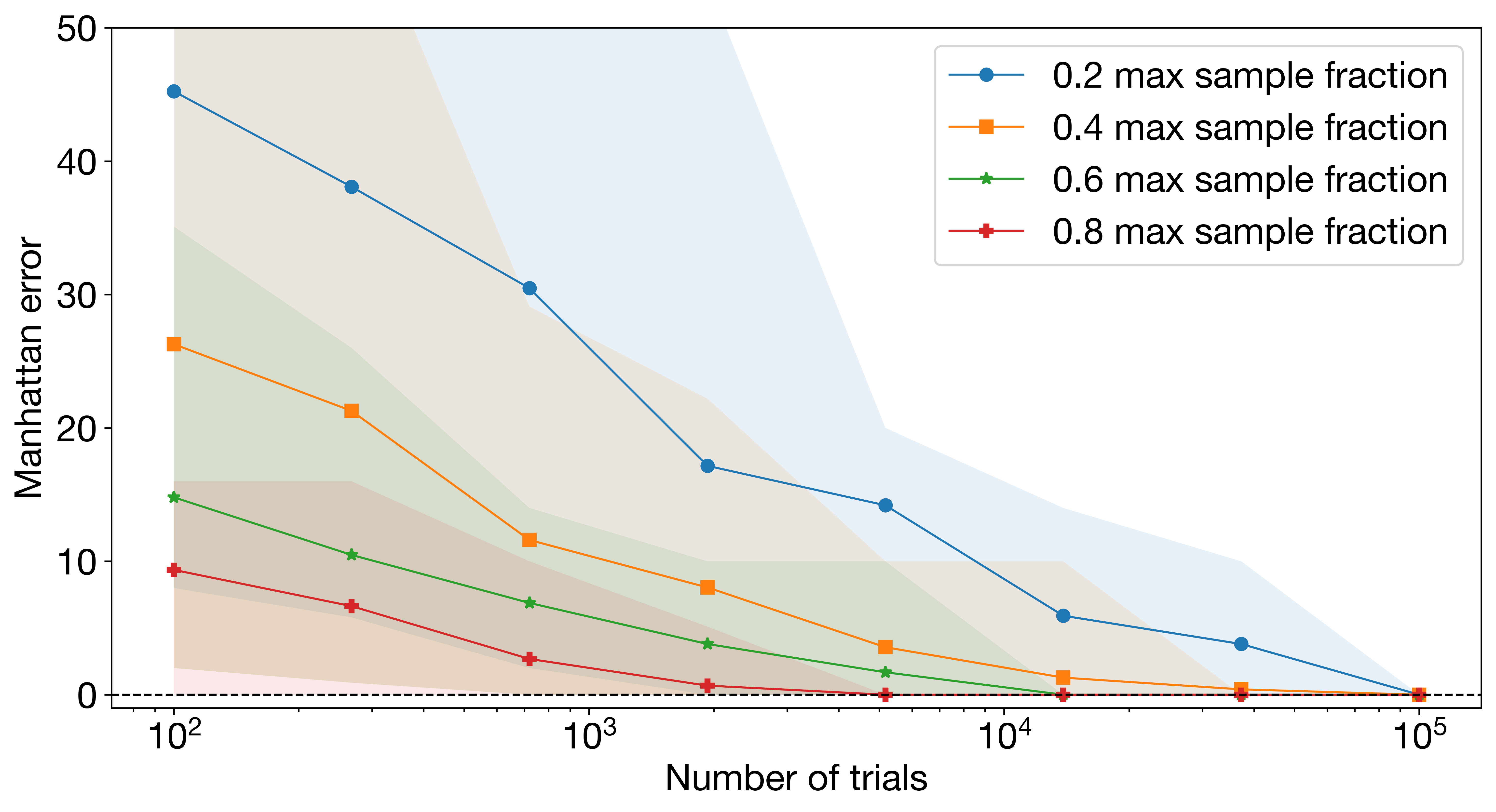}
         \caption{$\textbf{K=2}, N=100$ $(N_1=40, N_2=60)$}
         \label{fig:appendix_scaling_k=2}
     \end{subfigure}
     \hfill
     \begin{subfigure}[b]{0.49\textwidth}
         \centering
         \includegraphics[width=\textwidth]{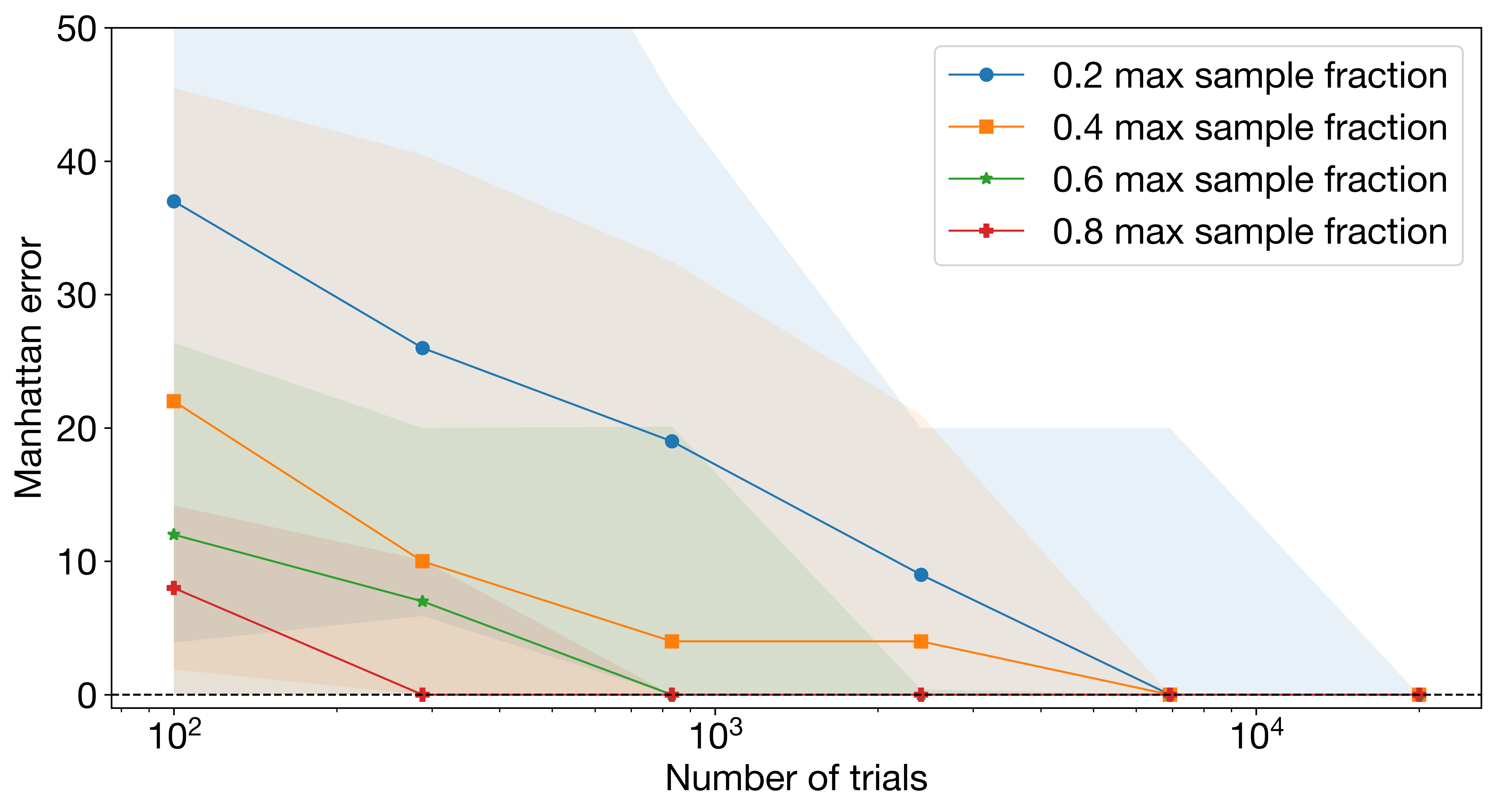}
         \caption{$\textbf{K=3}, N=100$ $(N_1=50, N_2=30, N_3=20)$}
         \label{fig:appendix_scaling_k=3}
     \end{subfigure}
     \caption{Maximum likelihood estimate Manhattan error for different numbers of observations at different max sample fractions ($N=100$). 90\% confidence interval over 50 random seeds.}
\end{figure}

\begin{figure}[h]
     \centering
     \begin{subfigure}[b]{0.49\textwidth}
         \centering
         \includegraphics[width=\textwidth]{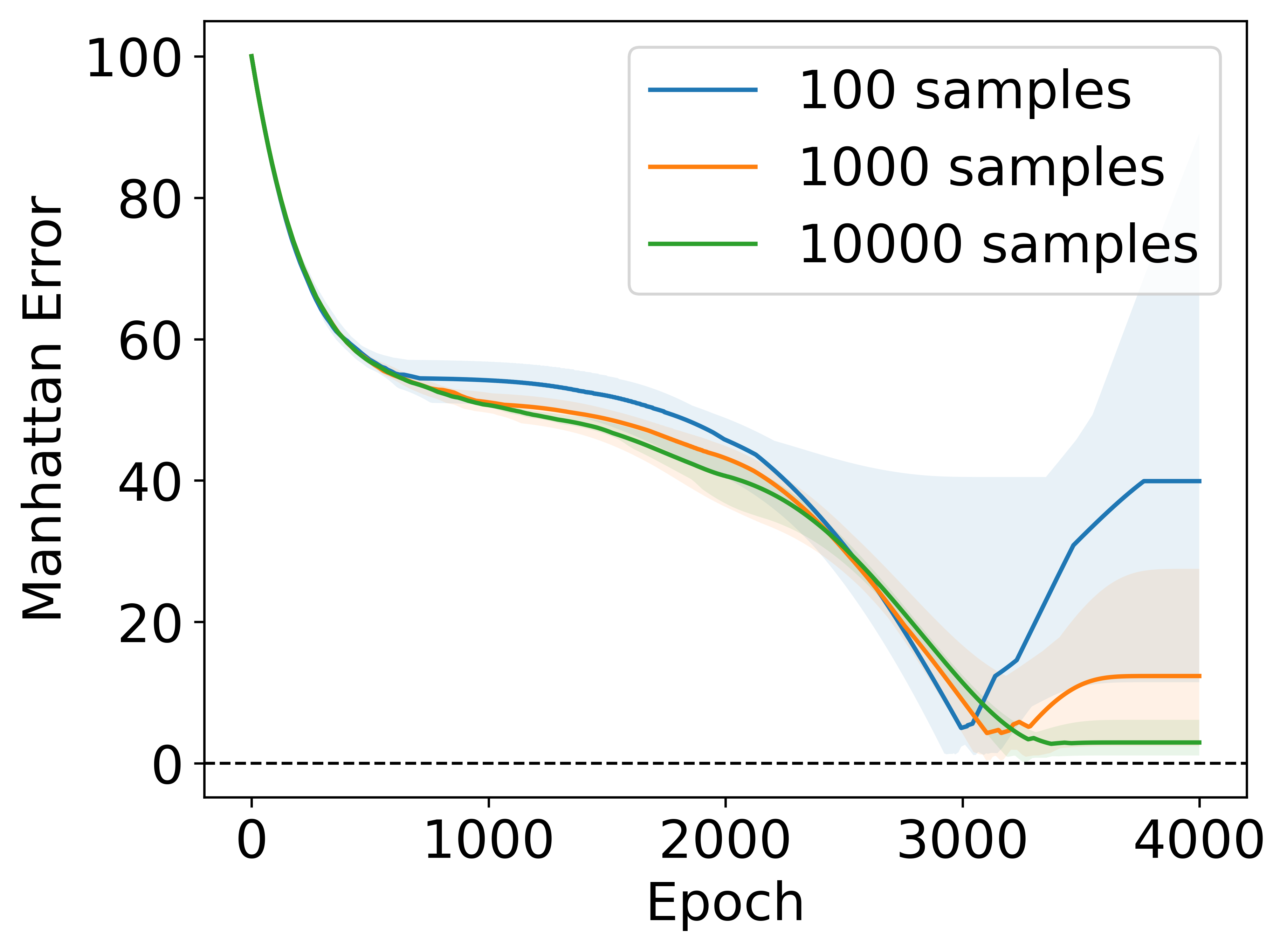}
         \caption{$\textbf{K=2}, N=100$ $(N_1=30, N_2=70)$}
         \label{fig:appendix_sgd_k=2}
     \end{subfigure}
     \hfill
     \begin{subfigure}[b]{0.49\textwidth}
         \centering
         \includegraphics[width=\textwidth]{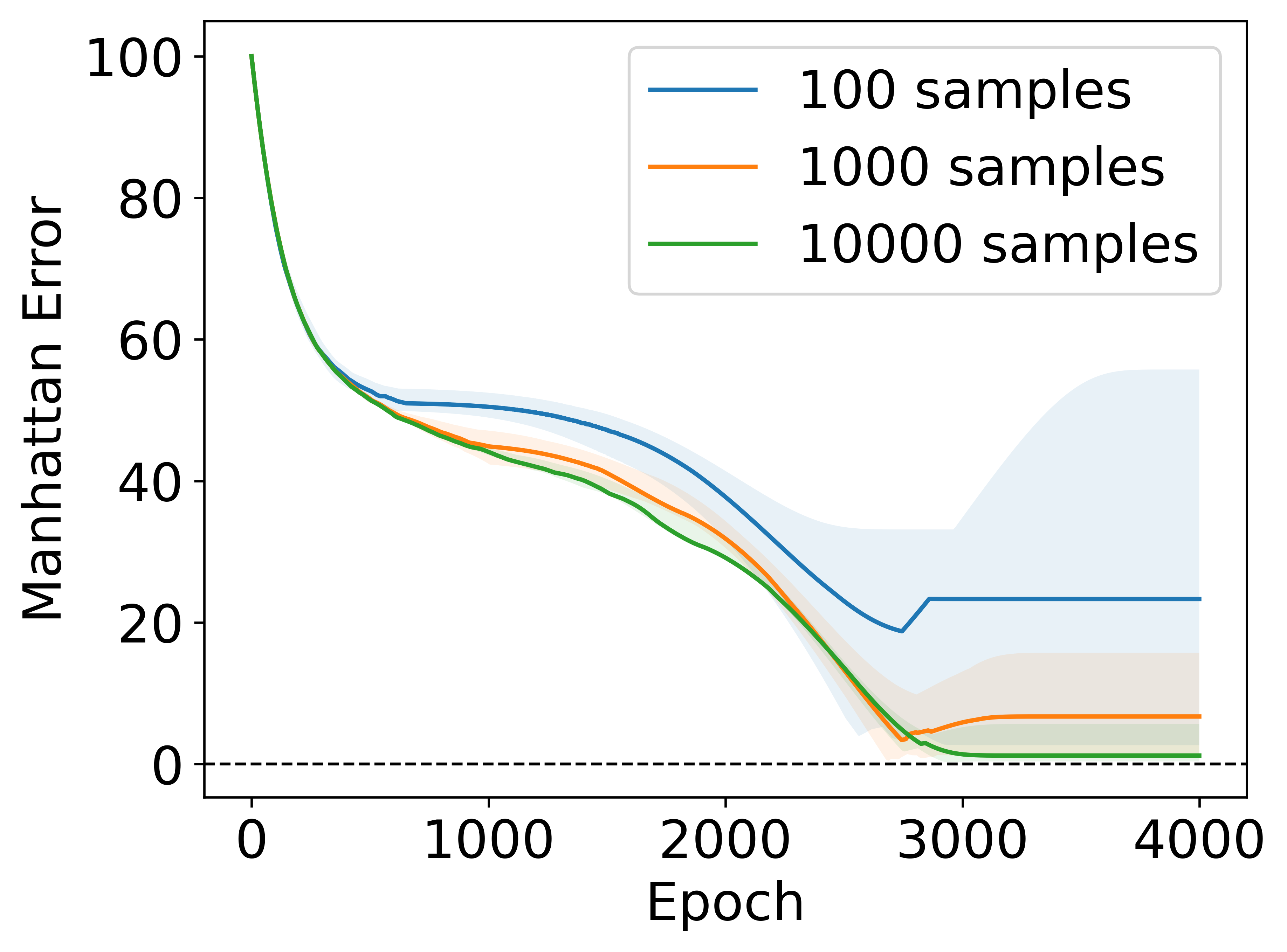}
         \caption{$\textbf{K=3}, N=100$ $(N_1=50, N_2=30, N_3=20)$}
         \label{fig:appendix_sgd_k=3}
     \end{subfigure}
     \caption{Maximum likelihood estimate Manhattan error per training epoch for different numbers of trials ($N=100$, max sample fraction 0.4). 90\% confidence interval over 20 random seeds.}
     \label{fig:appendix_sgd}
\end{figure}

\begin{figure}[h]
     \centering
     \begin{subfigure}[b]{0.32\textwidth}
         \centering
         \includegraphics[width=\textwidth]{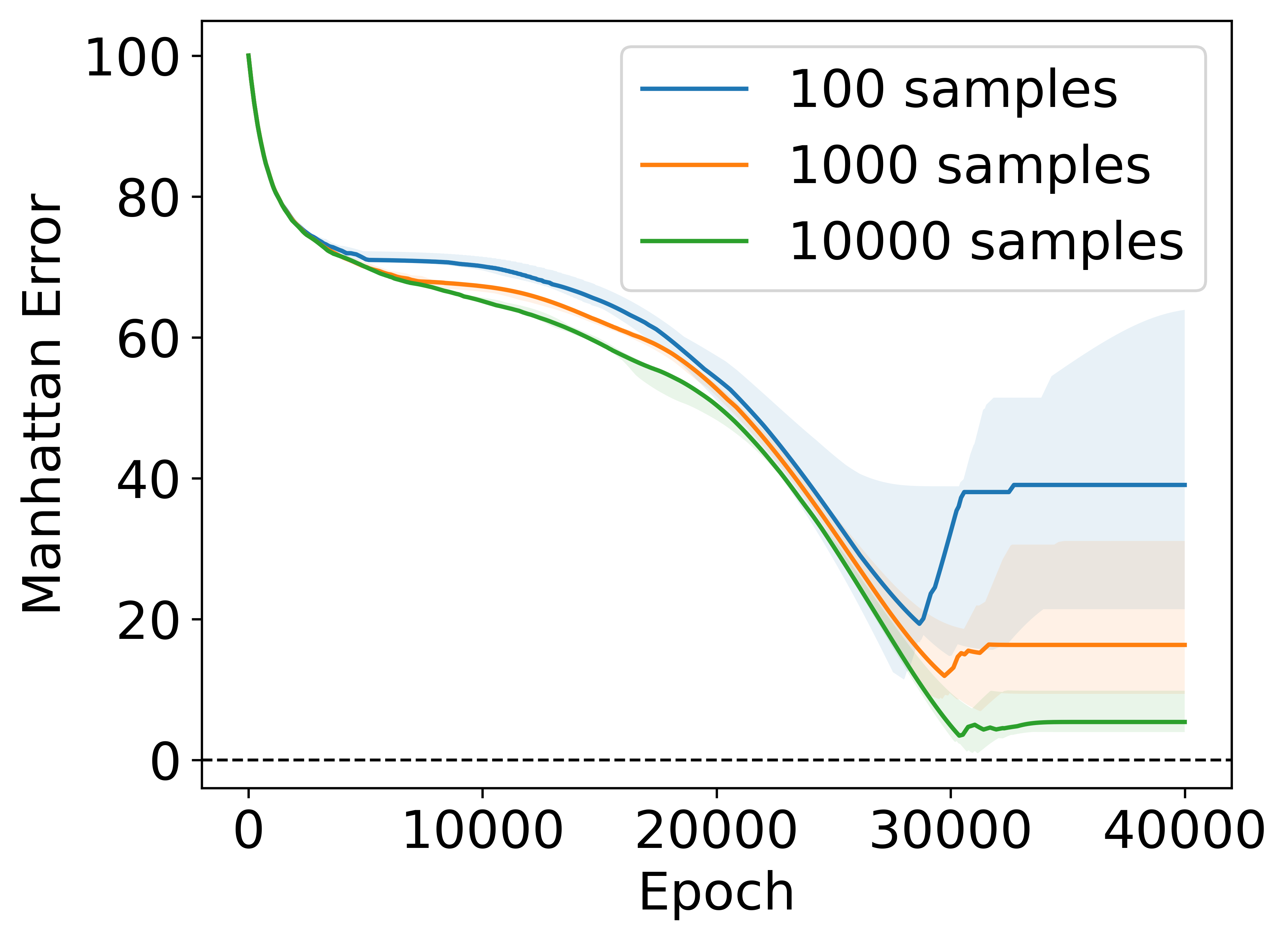}
         \caption{$f_{max}=0.2$}
         \label{fig:appendix_sgd_k=3_0.2frac}
     \end{subfigure}
     \hfill
     \begin{subfigure}[b]{0.32\textwidth}
         \centering
         \includegraphics[width=\textwidth]{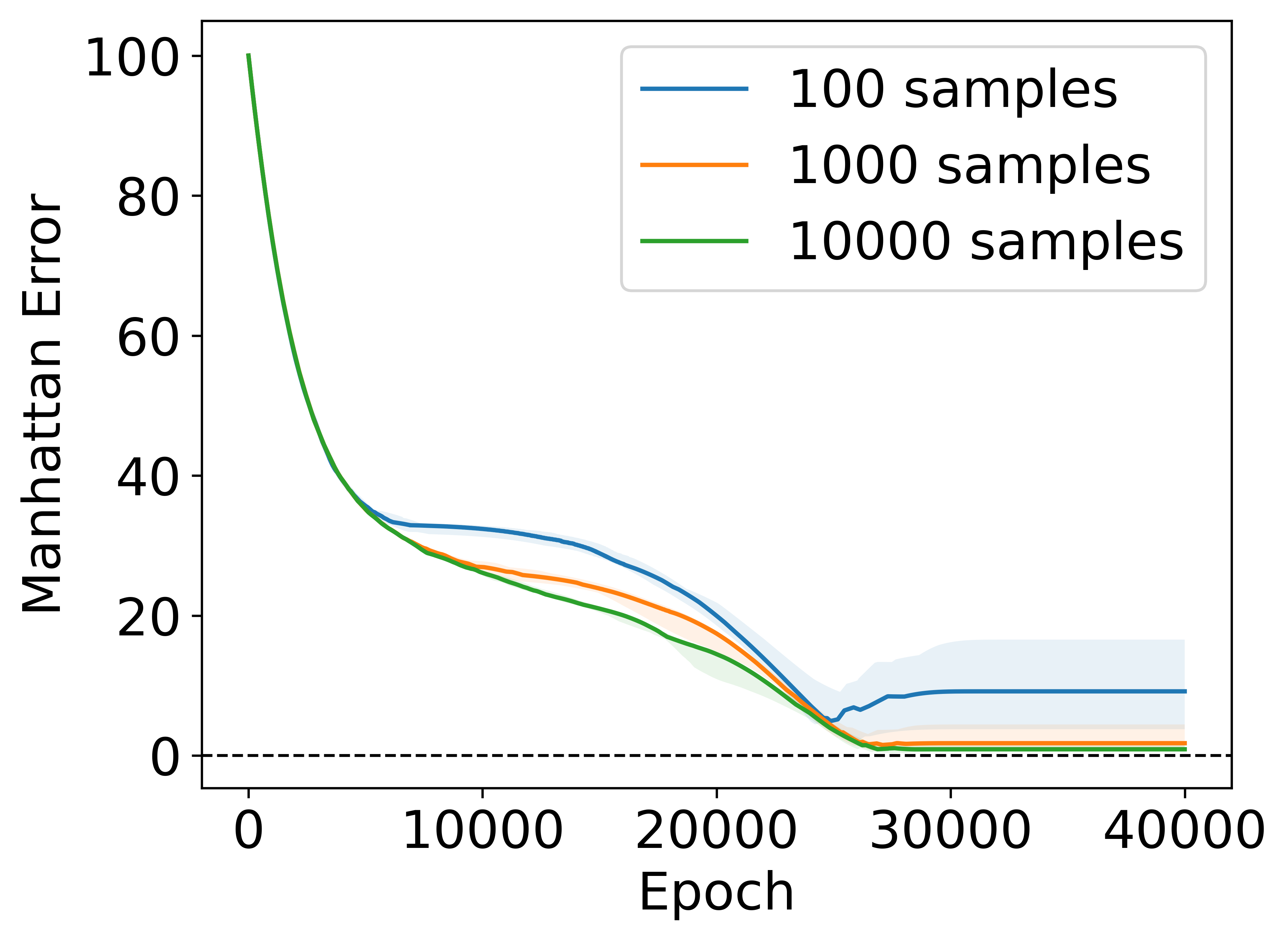}
         \caption{$f_{max}=0.6$}
         \label{fig:appendix_sgd_k=3_0.6frac}
     \end{subfigure}
     \hfill
     \begin{subfigure}[b]{0.32\textwidth}
         \centering
         \includegraphics[width=\textwidth]{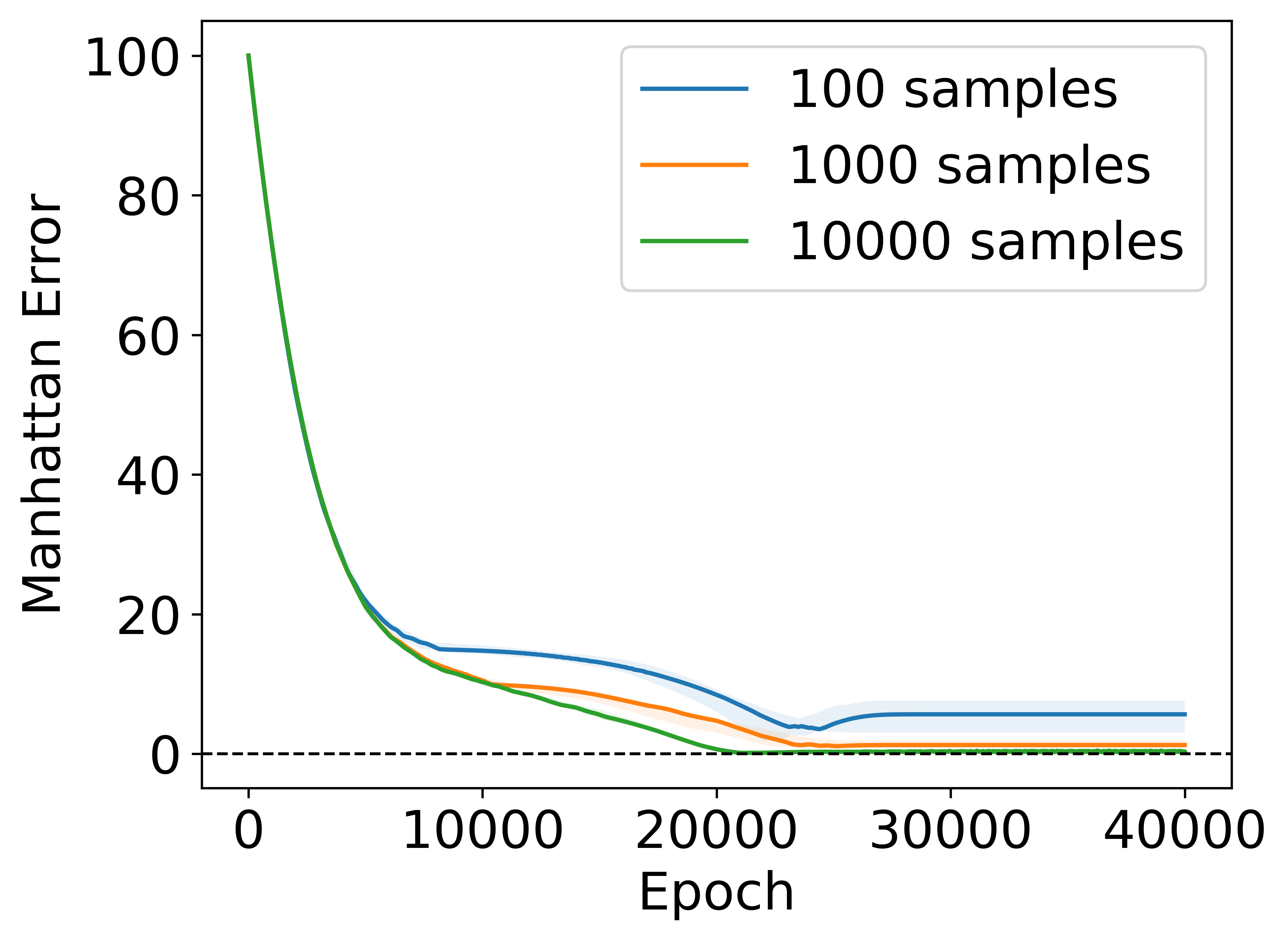}
         \caption{$f_{max}=0.8$}
         \label{fig:appendix_sgd_k=3_0.8frac}
     \end{subfigure}
     \caption{Maximum likelihood estimate Manhattan error per training epoch for different numbers of trials and different $f_{max}$ ($K=3, N=100, N_1=50, N_2=30, N_3=20$). 50\% confidence interval over 20 random seeds.}
     \label{fig:appendix_sgd2}
\end{figure}

\end{document}